\documentclass[a4paper,fleqn]{cas-dc}

\usepackage[numbers,sort&compress]{natbib}

\usepackage{graphicx}
\usepackage{amsmath}
\usepackage{amssymb}
\usepackage{booktabs}
\usepackage{multirow}
\usepackage{tabularx}
\usepackage{makecell}
\usepackage{siunitx}
\usepackage{xcolor}
\usepackage{microtype}
\usepackage{placeins}
\usepackage{listings}
\usepackage[most]{tcolorbox}
\usepackage{arydshln}
\usepackage[table]{xcolor}
\usepackage{seqsplit}
\usepackage{pdfpages}
\usepackage{pax}

\newcommand{\best}[1]{\textbf{#1}}
\newcommand{\negdelta}[1]{\textcolor{red!70!black}{#1}}
\newcommand{\codeid}[1]{\nolinkurl{#1}}

\begin{document}

\let\WriteBookmarks\relax

\setcounter{topnumber}{4}
\setcounter{bottomnumber}{2}
\setcounter{totalnumber}{6}
\setcounter{dbltopnumber}{3}
\renewcommand{\topfraction}{0.95}
\renewcommand{\bottomfraction}{0.80}
\renewcommand{\textfraction}{0.05}
\renewcommand{\floatpagefraction}{0.80}
\renewcommand{\dbltopfraction}{0.95}
\renewcommand{\dblfloatpagefraction}{0.80}
\setlength{\textfloatsep}{10pt plus 2pt minus 2pt}
\setlength{\floatsep}{8pt plus 2pt minus 2pt}
\setlength{\dbltextfloatsep}{10pt plus 2pt minus 2pt}
\setlength{\dblfloatsep}{8pt plus 2pt minus 2pt}
\setlength{\emergencystretch}{1em}

\shorttitle{Large-Scale Biomedical RAG: Corpus Scaling, Document Representation, and Efficiency Trade-offs}
\shortauthors{Sánchez-Freire et al.}

\title[mode=title]{Large-Scale Biomedical RAG: Corpus Scaling, Document Representation, and Efficiency Trade-offs}

\author[1]{Adrián Sánchez-Freire}
\cormark[1]
\ead{adrian.sanchez@ehu.eus}

\author[1]{Gonzalo Pérez-Gómez}
\author[1]{Aitziber Atutxa}
\author[1]{Ander Barrena}

\affiliation[1]{
  organization={HiTZ Basque Center for Language Technology, University of the Basque Country (EHU)},
  addressline={P.º Rafael Moreno "Pitxitxi", 2},
  city={Bilbao},
  postcode={48013},
  state={Biscay},
  country={Spain}
}

\cortext[cor1]{Corresponding author}
\begin{abstract}
We present a controlled, large-scale evaluation of how corpus scale, document
representation, and reranking affect the effectiveness and efficiency of
biomedical Retrieval-Augmented Generation (RAG) systems. Retrieval collections
were progressively expanded while retaining the relevant source documents,
enabling consistent comparisons across corpus sizes. We compared abstract,
flat full-text, and hierarchical full-text representations across first-stage
retrieval, reranking, and end-to-end multiple-choice question answering.
Configurations selected on development data were evaluated on five external
biomedical benchmarks and a held-out test set, while retrieval latency and
index storage were measured to quantify operational cost. Performance patterns
observed with small or controlled collections did not consistently persist as
collections grew, and the usefulness of reranking depended on the corpus and
retrieval setting. RAG improved three of four model configurations across the
external benchmarks, with mean gains of up to 9.6 percentage points, and
improved all four configurations on the held-out test set by 10.2--30.3 points.
Abstract retrieval performed best when reference evidence was abstract-aligned,
whereas full-text representations were generally more effective when richer
evidence was needed. Hierarchical representation matched or outperformed flat
full-text retrieval in most end-to-end comparisons while requiring 67\% fewer
indexed units and 44--56\% less storage. These findings show that corpus scale
should be treated as an explicit evaluation variable in biomedical RAG, and
that the most appropriate design depends on evidence format, model behavior,
and operational constraints.
\end{abstract}


\let\printorcid\relax

\maketitle


\section{Introduction}
\label{sec:introduction}

Large language models (LLMs) are increasingly used for biomedical and medical
question answering, but answers based only on parametric knowledge may be
incomplete, outdated, or difficult to verify. Retrieval-Augmented Generation
(RAG) addresses this limitation by providing external documents at inference
time, allowing the generator to condition its answer on explicit evidence
rather than only on information encoded in its parameters
\citep{lewis2021rag}.

The effectiveness of biomedical RAG depends not only on the generator, but also
on the retrieval corpus and on how its documents are represented. Expanding the
corpus may increase the availability of relevant evidence, but it also
introduces additional distractors and raises retrieval and storage costs.
Document representation creates a related trade-off. Abstracts are compact and
efficient, but may omit answer-relevant details. Full-text retrieval preserves
richer evidence, but substantially increases the number of indexed units and
may fragment information across independent chunks. Hierarchical
representations provide an alternative by retaining relationships between
different parts of a document.

These factors cannot be considered independently. As the source corpus grows,
each representation produces a retrieval index of a different size and
structure, which may affect first-stage retrieval, reranking, the evidence
provided to the generator, and ultimately the quality of the generated answer.
A representation that improves evidence coverage may therefore be impractical
because of its storage or latency requirements, while a reranker that improves
the final ranking may not necessarily improve downstream answer accuracy.

This study investigates these interactions through a controlled evaluation of
biomedical RAG systems under increasing corpus scale. We distinguish the number
of source documents from the representation-dependent number of retrievable
units and compare three document representations: abstracts, flat full-text
chunks, and TreeRAG-style hierarchical representations
\citep{treerag}. The experiments use PubMed as a large collection of real
biomedical literature and MIRIAD~\citep{miriad} as a controlled setting with
question-linked source passages.

We evaluate the RAG pipeline both modularly and end-to-end. First-stage
retrieval and reranking are assessed separately to determine how corpus growth
affects evidence selection. Downstream performance is evaluated through
multiple-choice question answering (MCQA), using development datasets, external
biomedical benchmarks, and a held-out MIRIAD test split. We compare LLM-only and
RAG configurations across compact general-purpose, biomedical, and
reasoning-oriented generators, and use a dataset-specific Oracle condition to
characterize generation when reference evidence is directly available.
Retrieval latency and index storage are also measured to quantify the practical
cost of each configuration.

To the best of our knowledge, existing biomedical RAG studies commonly evaluate
relatively small or task-specific retrieval collections, while a smaller number
use large indexes. However, these systems are generally assessed at a single
fixed corpus scale, and corpus growth is rarely analyzed jointly with document
representation, retrieval, reranking, downstream generation, and efficiency.
Consequently, it remains unclear whether conclusions obtained at one collection
size remain valid as biomedical knowledge resources grow. This study addresses
this gap through a controlled, system-level empirical evaluation of biomedical
RAG under corpus growth.

This work makes the following contributions:

\begin{itemize}

\item It introduces a controlled corpus-scaling protocol that retains gold
sources while progressively adding non-gold documents and separates
source-document scale from representation-dependent retrieval-index scale.

\item It conducts an end-to-end evaluation linking corpus growth and document
representation to first-stage retrieval, reranking, downstream MCQA accuracy,
latency, and storage.

\item It provides empirical evidence that advantages observed at smaller
corpus scales do not consistently persist as the retrieval collection grows,
demonstrating the limitations of evaluating biomedical RAG at a single fixed
scale.

\item It derives empirically grounded guidance for selecting biomedical RAG
configurations according to evidence characteristics, corpus scale, and
operational constraints.

\end{itemize}

Together, these contributions characterize the robustness and operational
trade-offs of biomedical RAG configurations under controlled corpus growth
rather than at an isolated collection size. The resulting evidence is intended to support researchers and developers
designing scalable, literature-grounded RAG systems for biomedical
applications.


\section{Related Work}
\label{sec:related}

\subsection{Retrieval-Augmented Biomedical Question Answering}
\label{sec:related-biomedical-rag}

Retrieval-Augmented Generation (RAG) conditions language models on information
retrieved from external sources for knowledge-intensive
tasks~\citep{lewis2021rag}. Subsequent work has explored large retrieval
memories, in-context and active retrieval, and modular retrieval--generation
architectures~\citep{borgeaud2022retro,ram2023context,
jiang2023active,mialon2023augmented}. Surveys describe this broader design
space and its role in grounding outputs and incorporating external
knowledge~\citep{gao2023rag_survey,zhao2024rag_survey}.

Biomedical question answering has evolved from task-specific and
retrieval-based systems~\citep{jacquemart2003medicalqa,athenikos2010,
jin2022survey} to domain-pretrained encoders~\citep{lee2020biobert,
gu2021domain,yasunaga2022linkbert} and general-purpose or medically specialized
LLMs~\citep{singhal2023medpalm,singhal2023medpalm2,
nori2023capabilities}. Medical QA benchmarks, including those consolidated in
MIRAGE~\citep{medrag}, provide a standardized multiple-choice setting for
evaluating how retrieval choices affect final answer accuracy.

\subsection{Document Representation in RAG}
\label{sec:related-document-representation}

Passage-level retrieval makes search over large collections tractable and
allows generators to use multiple evidence units
\citep{karpukhin2020dpr,izacard2021fid}. However, fixed-size segmentation can
split relevant evidence, remove document-level context, and create redundant
units. Retrieval granularity therefore affects both retrieval and downstream
generation~\citep{liu2021dense_hierarchical,
bhat2025rethinking_chunk_size}.

This issue is especially relevant for scientific literature. Abstracts provide
compact summaries, whereas full-text articles may contain evidence absent from
the abstract~\citep{dasigi2021qasper}. Hierarchical and long-document methods
instead seek to preserve broader document
structure~\citep{sarthi2024raptor,jiang2024longrag}. TreeRAG organizes documents
into tree-structured retrieval units, enabling comparison between
abstract-level, flat full-text, and hierarchical
representations~\citep{treerag}.

\subsection{Corpus Scaling and Efficiency}
\label{sec:related-corpus-scaling}

Large RAG systems commonly use fixed retrieval collections ranging from
millions of passages to much larger external
memories~\citep{karpukhin2020dpr,izacard2021fid,borgeaud2022retro,
izacard2023atlas}. General-domain scaling studies show that larger collections
can improve evidence coverage, but with diminishing
returns~\citep{less_llm_more_documents}.

Biomedical retrieval-augmented systems have been evaluated across different QA
tasks, corpora, retrieval strategies, and evidence representations, including
the studies of Otegi et al.~\citep{otegi2022covidqa}, Xiong et
al.~\citep{medrag}, Alonso et al.~\citep{medexpqa}, Vladika and
Matthes~\citep{vladika2024healthqa}, Li et al.~\citep{li2025biomedrag}, Lopez et
al.~\citep{lopez2025clear}, and Sohn et al.~\citep{sohn2025rag2}. Stuhlmann et
al.~\citep{efficient_biomed_rag} additionally examine retrieval efficiency over
a large PubMed-derived collection. Context-selection studies further show that
additional retrieved material may introduce distractors or unfavorable
retrieval--generation
trade-offs~\citep{optimal_search_rag,dynamic_context_selection}.

Table~\ref{tab:sota_comparison} summarizes these lines of work. Existing studies investigate individual aspects of corpus scale, document representation, ranking, and efficiency, but do not jointly analyze their interaction under controlled biomedical corpus growth. To the best of our knowledge, this work is the first to unify these dimensions within a single experimental framework, enabling a systematic study of the trade-offs among corpus scale, document representation, retrieval performance, and computational efficiency.


\begin{table*}[!t]
\centering
\footnotesize
\setlength{\tabcolsep}{4pt}
\renewcommand{\arraystretch}{1.1}

\begin{tabular}{@{}llcccc@{}}
\toprule
Reference & Max. index & Biomed. & Scaling & Doc. repr. & Efficiency \\
\midrule

Borgeaud et al.~\citep{borgeaud2022retro}
& 28B units & -- & $\sim$ & -- & -- \\

Izacard et al.~\citep{izacard2023atlas}
& 387M units & -- & -- & -- & -- \\

Ning et al.~\citep{less_llm_more_documents}
& 264M units & -- & $\checkmark$ & -- & $\checkmark$ \\

Tao et al.~\citep{treerag}
& n.r. & $\sim$ & -- & $\checkmark$ & -- \\

\midrule

Otegi et al.~\citep{otegi2022covidqa}
& 400K units & $\checkmark$ & -- & $\checkmark$ & -- \\

Xiong et al.~\citep{medrag}
& 54.2M units & $\checkmark$ & -- & -- & -- \\

Alonso et al.~\citep{medexpqa}
& 54.2M units & $\checkmark$ & -- & -- & -- \\

Vladika and Matthes~\citep{vladika2024healthqa}
& 20.6M units & $\checkmark$ & -- & -- & -- \\

Stuhlmann et al.~\citep{efficient_biomed_rag}
& 24M units & $\checkmark$ & -- & -- & $\checkmark$ \\

Li et al.~\citep{li2025biomedrag}
& n.r. & $\checkmark$ & -- & $\sim$ & -- \\

Lopez et al.~\citep{lopez2025clear}
& 20K units & $\checkmark$ & -- & $\checkmark$ & $\checkmark$ \\

Sohn et al.~\citep{sohn2025rag2}
& 116.7M units & $\checkmark$ & -- & -- & -- \\

\midrule

\textbf{This work}
& \textbf{117.4M units}
& $\checkmark$
& $\checkmark$
& $\checkmark$
& $\checkmark$ \\

\bottomrule
\end{tabular}

\caption{Comparison with representative retrieval-augmented studies.
\textit{Max. index} reports the largest retrieval collection evaluated by each
study in its reported indexing unit. The remaining columns indicate biomedical
or medical evaluation, controlled corpus scaling, document-representation
comparison, and efficiency analysis. $\checkmark$ denotes explicit coverage,
$\sim$ partial or indirect coverage, -- not addressed, and n.r. not reported.}
\label{tab:sota_comparison}
\end{table*}


\section{Methods}
\label{sec:methods}

\subsection{Datasets and Task Format}
\label{sec:datasets-task-format}

We formulate all tasks as multiple-choice question answering (MCQA), enabling
consistent evaluation through accuracy across datasets and experimental
configurations.

Development experiments use two datasets. \textbf{MedAESQA} is a clinically
grounded QA dataset containing 40 expert-curated questions supported by
biomedical evidence~\citep{medaesqa}. The complete dataset is used for
development.
\textbf{MIRIAD} is a large synthetic biomedical QA dataset containing
automatically generated question--answer pairs linked to source
passages~\citep{miriad}. We sample 1,000 instances and divide them into 200 development and
800 test examples.\footnote{The sampling script is available in the main experimental repository; see the Data Availability section.}

MedAESQA and MIRIAD are originally formulated as open-ended QA tasks. We
convert them to MCQA while preserving the original reference answer as the
correct option and generating three incorrect distractors. The conversion
pipeline consists of three stages: generating medically plausible candidates,
rewriting them to reduce superficial cues such as differences in length or
syntax, and filtering and ranking the candidates to select the final
distractors. Correct-answer positions are balanced across option labels.
Implementation details, including model configuration, decoding parameters,
and filtering heuristics, are provided in Supplementary Section B.

Final evaluation additionally uses MMLU-Med~\citep{mmlu},
MedQA-US~\citep{medqa}, MedMCQA~\citep{medmcqa},
PubMedQA~\citep{pubmedqa}, and BioASQ yes/no~\citep{bioasq}. We use the
preprocessed versions introduced in MIRAGE~\citep{medrag}, with minor
modifications to align them with our evaluation pipeline. MedAESQA and the
200-instance MIRIAD development split are used for model selection and
configuration decisions. The external benchmarks and the remaining 800 MIRIAD
instances are used exclusively for testing.

\subsection{Corpora and Scaling Protocol}
\label{sec:corpora-scaling}

Retrieval experiments use two document collections. PubMed provides the
retrieval corpus for MedAESQA and the external biomedical benchmarks, whereas
MIRIAD source passages are used exclusively for MIRIAD experiments.

We construct corpus scales by progressively increasing the number of non-gold
documents in the retrieval pool while retaining the available gold sources.
Five shared scaling factors are applied independently to PubMed and MIRIAD:
\textit{control}, \textit{tiny}, \textit{small}, \textit{medium}, and
\textit{large}. The absolute corpus sizes differ, but the shared factors provide
corresponding corpus-growth conditions across the two collections. The
\textit{large} scale corresponds to the complete MIRIAD corpus. Because PubMed
can be expanded further, we include an additional \textit{full} setting
containing approximately 4.3 million source documents.

The scaling protocol is defined at the source-document level. However, the
number of units available to the retriever depends on how each document is
represented. Table~\ref{tab:corpus-index-scaling} therefore reports both source
documents and the resulting representation-dependent index sizes.

\subsection{Document Representations}
\label{sec:document-representations}

We compare three document representations:

\begin{itemize}
    \item \textbf{Abstract-level representation.}
    Each source document is represented by a single compact retrieval unit.
    \item \textbf{Full-text chunking.} Source documents are divided into chunks of up to 1,600 characters with a 200-character overlap using a recursive character-based splitter.
    \item \textbf{Tree-based representation.}
    Source documents are organized into hierarchical retrieval units using
    TreeRAG~\citep{treerag}.
\end{itemize}

For PubMed, all three representations are constructed from the same set of
publicly available full-text articles. The abstract-level representation uses
the original abstract associated with each article, whereas the full-text
content is used to construct the flat chunks and TreeRAG representations.

For MIRIAD, the original source passages serve as the full-text documents.
Because MIRIAD does not provide separate abstracts, we generate
LLM-based pseudo-abstracts from these passages. Passages containing at most
300 tokens are kept unchanged because they already match the intended

abstract-level granularity. Full-text chunks and TreeRAG units are constructed
from the original passages. Further details on pseudo-abstract generation and
TreeRAG construction are provided in Supplementary Sections C and D.

\begin{table*}[!t]
\centering
\footnotesize
\setlength{\tabcolsep}{4pt}
\renewcommand{\arraystretch}{1.1}

\begin{tabular}{@{}llrrrrr@{}}
\toprule
& & & & \multicolumn{3}{c}{\textbf{Retrieval units}} \\
\cmidrule(lr){5-7}

\textbf{Corpus}
& \textbf{Scale}
& \textbf{Factor ($\times$)}
& \textbf{Source docs}
& Abstract
& Full-text
& TreeRAG \\
\midrule

\multirow[t]{6}{*}{PubMed}
& Control & 1.0    & 0.3K   & 0.3K   & 11.2K   & 2.6K \\
& Tiny    & 6.6    & 1.8K   & 1.8K   & 53.4K   & 15.9K \\
& Small   & 44.1   & 11.8K  & 11.8K  & 330.6K  & 105.7K \\
& Medium  & 292.6  & 78.4K  & 78.4K  & 2.2M    & 700.2K \\
& Large   & 1942.2 & 520.5K & 520.5K & 14.4M   & 5.2M \\
& Full    & --     & 4.3M   & 4.3M   & 117.4M  & 38.3M \\
\midrule

\multirow[t]{5}{*}{MIRIAD}
& Control & 1.0    & 1.0K   & 1.0K   & 4.2K    & 4.2K \\
& Tiny    & 6.6    & 6.6K   & 6.6K   & 28.1K   & 27.4K \\
& Small   & 44.1   & 44.1K  & 44.1K  & 186.9K  & 182.2K \\
& Medium  & 292.6  & 292.6K & 292.6K & 1.2M    & 1.2M \\
& Large   & 1942.2 & 1.9M   & 1.9M   & 8.2M    & 8.0M \\
\bottomrule
\end{tabular}

\caption{Corpus scales and representation-dependent retrieval-index sizes.
\textit{Source docs} denotes the number of documents in each retrieval pool,
whereas \textit{Retrieval units} reports the number of indexed abstracts,
full-text chunks, or TreeRAG units. Factors are defined relative to the
corpus-specific control setting and applied independently to both corpora.
Values are abbreviated for readability; K denotes thousands and M denotes
millions. The \textit{full} setting is available only for PubMed.}
\label{tab:corpus-index-scaling}
\end{table*}

\subsection{Models}
\label{sec:models}

We evaluate models with different architectures and parameter scales at each
stage of the RAG pipeline. First-stage retrieval uses BM25~\citep{bm25} as a
sparse baseline, BGE-base-en-v1.5~\citep{bge_base_emb} as a lightweight dense
retriever with approximately 0.1B parameters, and
Qwen3-Embedding-8B~\citep{qwen3_emb} as a larger dense retriever with 8B
parameters.

For reranking, we compare Jina-Reranker-v3~\citep{jina_rerank}, a 0.6B
listwise reranker with a decoder-only backbone; BGE-Reranker-v2-M3~\citep{bge_m3_rerank1,bge_m3_rerank2},
a 0.6B cross-encoder; and Qwen3-Reranker-8B~\citep{qwen3_emb}, an 8B
LLM-based reranker. This selection covers different reranking architectures
and provides a comparison between compact and higher-capacity models.

Generation experiments use compact 4B-parameter models, reflecting growing
interest in smaller models whose computational requirements are more compatible
with deployment in biomedical and healthcare systems. We evaluate three models
in four configurations. Gemma-3-4B~\citep{gemma3} serves as a general-purpose
baseline, while MedGemma-1.5-4B~\citep{medgemma1.5} provides a biomedical
counterpart from the same model family. Qwen3.5-4B~\citep{qwen3.5} is evaluated
with thinking mode disabled and enabled; we refer to these configurations as
Qwen3.5-4B and Qwen3.5-4B-Think, respectively. Because both configurations use
the same underlying model, their comparison isolates the effect of thinking
mode while keeping the architecture and parameter count fixed.

\subsection{Evaluation Protocol}
\label{sec:evaluation-protocol}

\subsubsection{Retrieval and Reranking}
\label{sec:evaluation-retrieval-reranking}

Retrieval quality is measured using Macro Recall@$k$, which computes recall
for each query and then averages across queries. This gives equal weight to
each question regardless of the number of associated gold documents. Gold
evidence is matched at the source-document level so that retrieval units
derived from the same document are treated consistently across
representations.

For MedAESQA, gold sources correspond to the PubMed articles supporting each
question, although the original evidence is provided at the abstract level.
For MIRIAD, gold sources correspond to the original full-text passages linked
to each question.

We report Recall@100 for first-stage retrieval. Each reranker is applied to the
top 100 retrieved candidates, and reranking quality is evaluated using
Recall@5.

\subsubsection{End-to-End Evaluation}
\label{sec:evaluation-end-to-end}

End-to-end performance is measured using MCQA accuracy. We compare three
configurations: \textit{LLM-only}, in which the generator receives only the
question and answer options; \textit{RAG}, in which it additionally receives
the top five evidence units produced by the retrieval pipeline; and
\textit{Oracle}, in which retrieval is bypassed and gold evidence is supplied
directly.

The retrieval configuration used for end-to-end evaluation is selected using
the development results and fixed before evaluation on the test sets.

The Oracle configuration is evaluated for MedAESQA and MIRIAD, for which
question-linked gold evidence is available. It uses the gold abstracts for
MedAESQA and the gold full-text passages for MIRIAD. This configuration
measures generation performance when reference evidence is available and helps
distinguish retrieval failures from generation limitations. Because the
supplied evidence follows the annotation format of each dataset, Oracle results
are interpreted as dataset-specific reference points rather than absolute
upper bounds.

\subsubsection{Retrieval Efficiency}
\label{sec:retrieval-efficiency-methods}

Retrieval efficiency is evaluated using query-time latency and index storage.
Latency is measured per question for the same first-stage retrieval and
reranking configurations used in the retrieval-quality evaluation. We report
first-stage retrieval latency separately from retrieval-plus-reranking latency
to quantify the additional cost of reranking the top 100 candidates.

Index storage is measured after indexing as the total Elasticsearch index
footprint. This captures the deployment cost introduced by
representation-dependent growth in the number of retrieval units. Because
absolute latency depends on hardware, software, and implementation details, we
interpret latency comparatively across configurations measured under the same
experimental conditions. The computational environment and software versions
are described in Supplementary Section A.


\section{Results}
\label{sec:results}

The experimental results are organized into sequential phases corresponding to
the main components of the RAG pipeline. Throughout the study, corpus scale and
document representation are treated as factors of analysis: Abstract,
Full-text, and TreeRAG are evaluated across the experimental phases rather than
reduced to a single selected representation. Development results are used 
to select the first-stage retriever and the reranking strategy applied in the
subsequent end-to-end experiments.

\subsection{Phase I: Retriever Evaluation}
\label{sec:results-retriever}

Phase~I compares the retrievers on the MedAESQA and MIRIAD development sets
using Macro Recall@100. Retriever selection is based on performance at the
largest available corpus scale for each dataset: \textit{full} PubMed for
MedAESQA and \textit{large} MIRIAD for the MIRIAD DEV split.

\begin{figure*}[!t]
\centering
\includegraphics[width=\textwidth]
{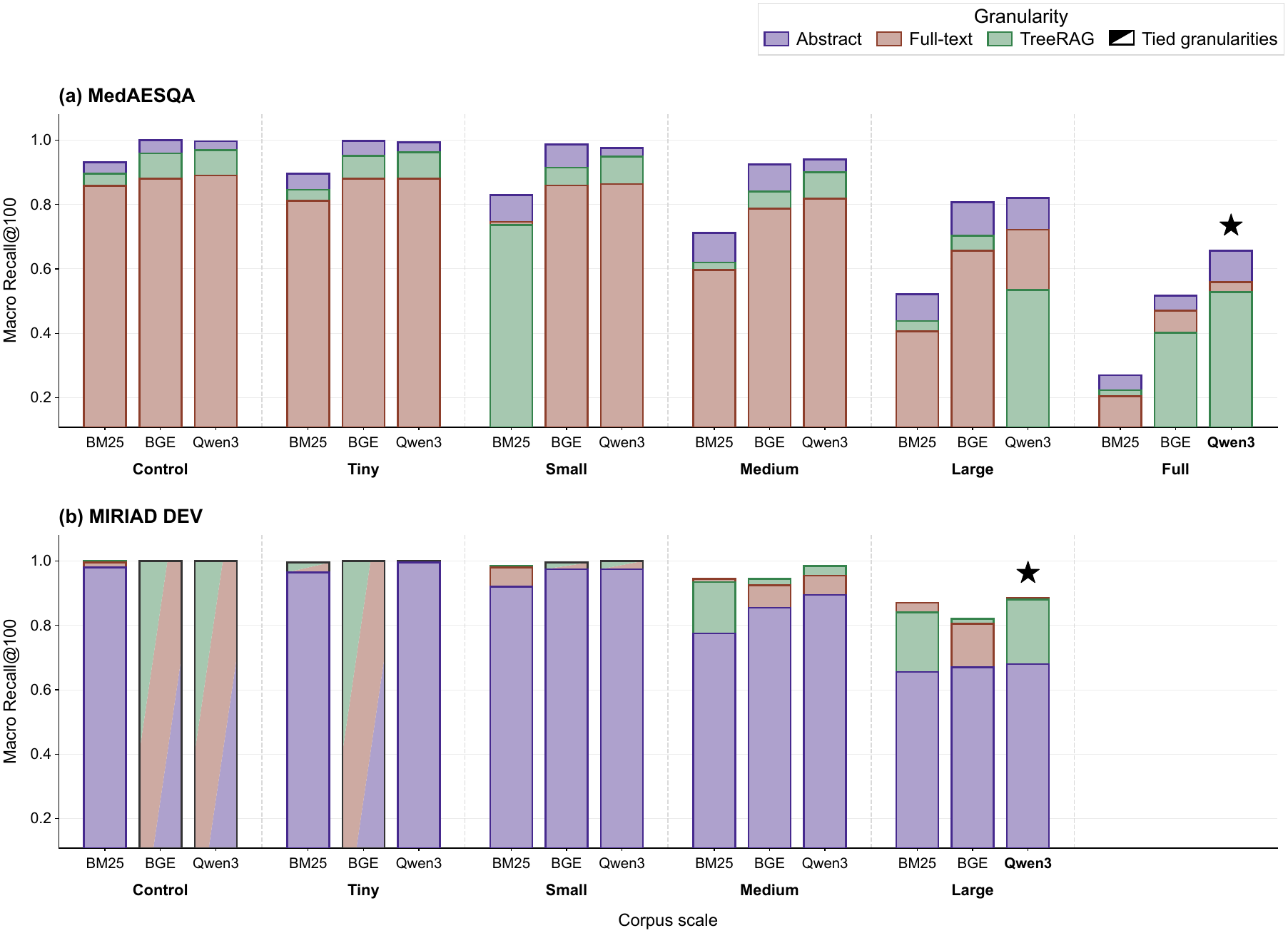}

\caption{Retriever evaluation on DEV using Macro Recall@100. Panel (a) shows
MedAESQA across PubMed corpus scales, and panel (b) shows MIRIAD DEV across
MIRIAD corpus scales. Colors denote the document representation: Abstract,
Full-text, or TreeRAG. Diagonal split fills indicate ties between
representations. Retriever labels correspond to BM25, BGE-base-en-v1.5 (BGE),
and Qwen3-Embedding-8B (Qwen3). The bold label and black star indicate the
highest-scoring retriever at the largest available scale for each dataset.}
\label{fig:phase1_retriever_eval}
\end{figure*}

Figure~\ref{fig:phase1_retriever_eval} reports retrieval performance across
corpus scales and document representations.

On MedAESQA, differences between retrievers become clearer as the PubMed corpus
grows. The Abstract representation performs best across most configurations,
consistent with the abstract-level evidence annotations in MedAESQA. At the
\textit{full} PubMed scale, Qwen3-Embedding-8B achieves the highest Macro
Recall@100, reaching 65.66\%. BGE-base-en-v1.5 remains competitive across most
scales but decreases more noticeably in the \textit{full} setting, while BM25
shows the strongest degradation as additional distractor documents are added.

On MIRIAD DEV, retrieval is close to saturation at the smallest scales, with
all retrievers obtaining high Macro Recall@100. Differences between models and
representations become more visible as the retrieval pool grows. The Abstract
representation is less effective than on MedAESQA. Unlike the original MIRIAD
passages used by the Full-text and TreeRAG representations, these abstracts are
synthetically generated pseudo-abstracts and may omit information relevant to
retrieval. At the \textit{large} MIRIAD scale, Qwen3-Embedding-8B achieves the
highest Macro Recall@100, reaching 88.50\%, while BM25 remains close at
87.00\%.

Qwen3-Embedding-8B obtains the highest Macro Recall@100 at the largest
available scale in both development datasets and is therefore selected as the
first-stage retriever for the subsequent phases.

\subsection{Phase II: Reranker Evaluation}
\label{sec:results-reranker}

Phase~II fixes Qwen3-Embedding-8B as the first-stage retriever and compares
No-Reranker, BGE-Reranker-v2-M3, Jina-Reranker-v3, and
Qwen3-Reranker-8B. In the No-Reranker condition, the top five evidence units
are taken directly from the first-stage ranking. Configurations are evaluated
using Macro Recall@5, with reranker selection based on \textit{full} PubMed for
MedAESQA and \textit{large} MIRIAD for MIRIAD DEV.

\begin{figure*}[!t]
\centering
\includegraphics[width=\textwidth]
{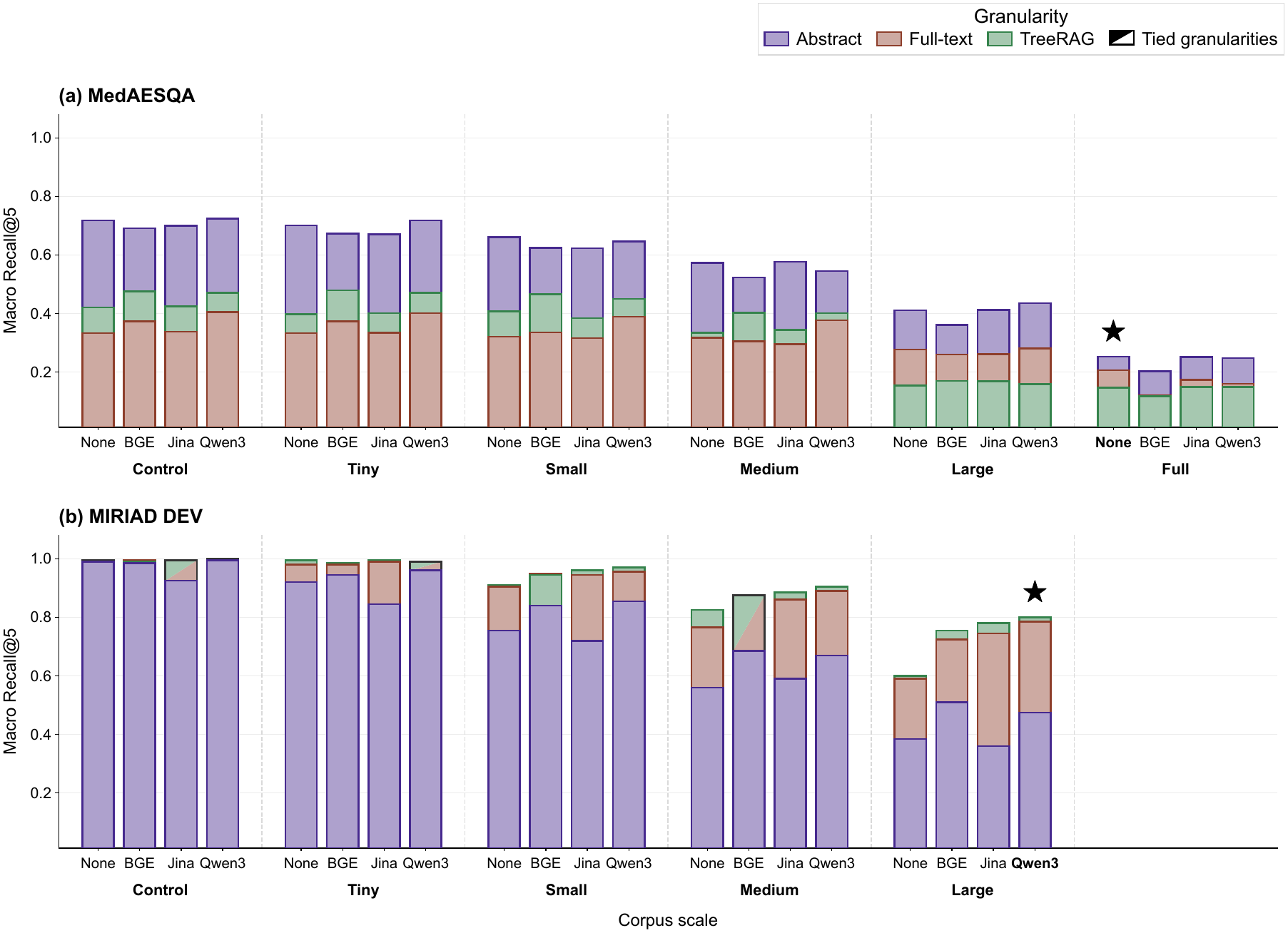}

\caption{Reranker evaluation on DEV using Macro Recall@5. Panel (a) shows
MedAESQA across PubMed corpus scales, and panel (b) shows MIRIAD DEV across
MIRIAD corpus scales. All configurations use Qwen3-Embedding-8B as the
first-stage retriever. Colors denote the document representation: Abstract,
Full-text, or TreeRAG. Diagonal split fills indicate ties between
representations. Labels correspond to No-Reranker, BGE-Reranker-v2-M3 (BGE),
Jina-Reranker-v3 (Jina), and Qwen3-Reranker-8B (Qwen3). The bold label and
black star indicate the highest-scoring reranking configuration at the largest
available scale for each dataset.}
\label{fig:phase2_reranker_eval}
\end{figure*}

Figure~\ref{fig:phase2_reranker_eval} reports the results across corpus scales
and document representations.

On MedAESQA, reranking does not provide a consistent improvement over the
first-stage ranking. No-Reranker remains comparable to the reranked
configurations across scales and obtains the highest Macro Recall@5 at the
\textit{full} PubMed scale, reaching 25.28\% with the Abstract representation.
Performance generally decreases as the PubMed corpus grows, while Abstract
remains the strongest representation across all configurations.

On MIRIAD DEV, the effect of reranking becomes clearer at larger corpus scales.
At the control and tiny scales, Macro Recall@5 is close to saturation for most
configurations. As the retrieval pool grows, the reranked configurations more
consistently outperform No-Reranker, particularly with the Full-text and
TreeRAG representations. At the \textit{large} MIRIAD scale,
Qwen3-Reranker-8B achieves the highest Macro Recall@5, reaching 80.00\% with
TreeRAG.

Reranker selection therefore differs by corpus. For subsequent PubMed-based
experiments, we retain the original Qwen3-Embedding-8B ranking without
reranking. For MIRIAD-based experiments, we apply Qwen3-Reranker-8B to the
first-stage results.

\subsection{Phase III: End-to-End Evaluation}
\label{sec:results-end-to-end}

Phase~III evaluates whether the retrieval configurations selected in the
previous phases improve downstream MCQA accuracy. All corpus scales and
document representations remain under evaluation. PubMed-based experiments
use Qwen3-Embedding-8B without reranking, whereas MIRIAD experiments use
Qwen3-Embedding-8B followed by Qwen3-Reranker-8B. We compare the generators
described in Section~\ref{sec:models} under the LLM-only, RAG, and Oracle
conditions.

These experiments use MedAESQA and MIRIAD DEV, which were also used to select
the retrieval components. The results therefore characterize development-stage
behavior rather than generalization to unseen test sets. Complete accuracy
values are reported in Supplementary Section E.

Figure~\ref{fig:phase3_e2e_eval} reports RAG accuracy gains, in percentage
points, relative to the corresponding LLM-only baseline for each generator.

\begin{figure*}[!t]
\centering

\includegraphics[
    width=\textwidth,
    height=0.82\textheight,
    keepaspectratio
]{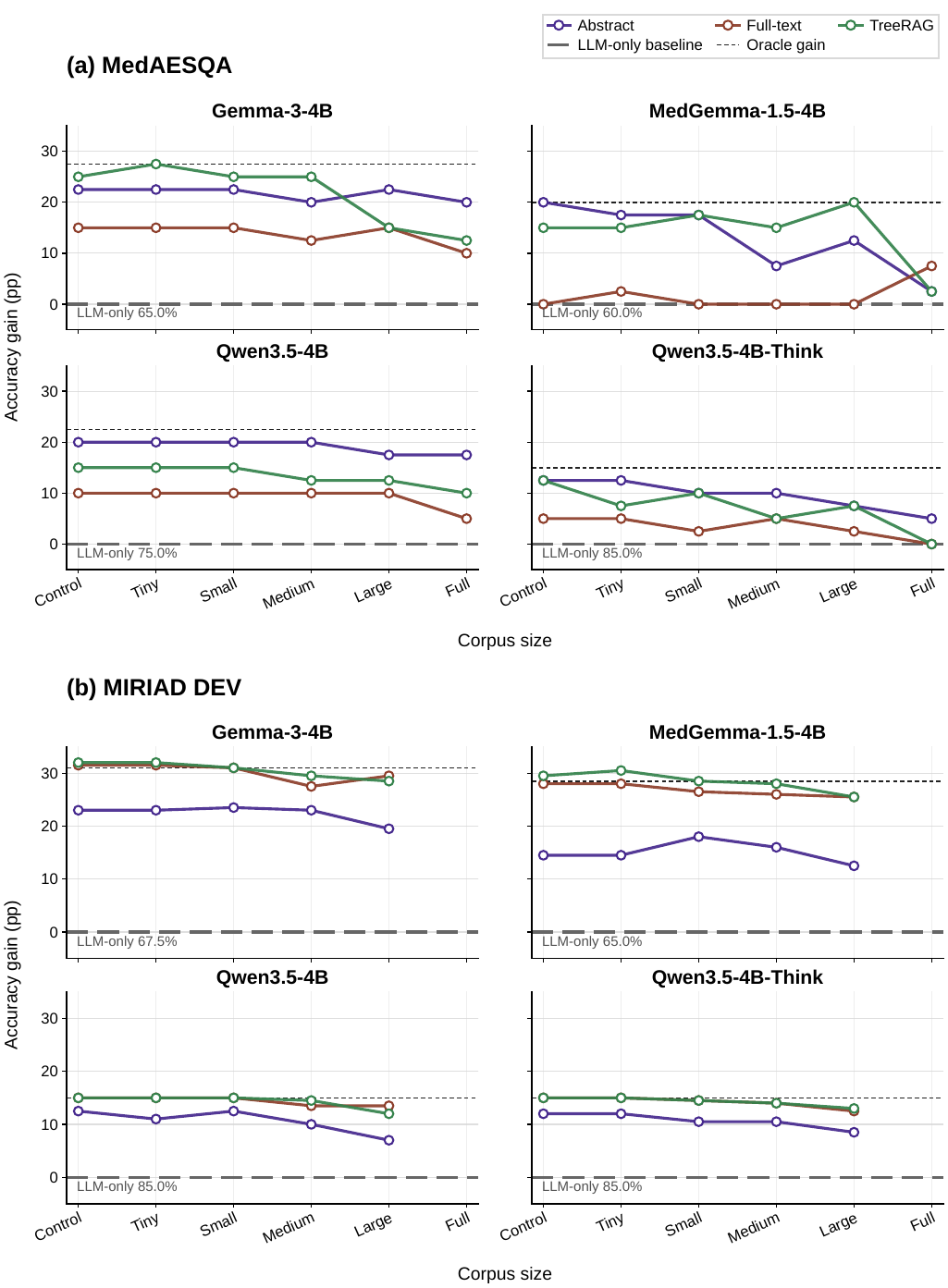}

\caption{End-to-end MCQA accuracy gains over each generator's LLM-only
baseline for (a) MedAESQA and (b) MIRIAD DEV, reported in percentage points
across corpus scales and document representations. Horizontal reference lines
indicate the LLM-only baseline at zero and the dataset-specific Oracle gain.
MedAESQA uses Qwen3-Embedding-8B without reranking, whereas MIRIAD DEV uses
Qwen3-Embedding-8B followed by Qwen3-Reranker-8B.}
\label{fig:phase3_e2e_eval}
\end{figure*}

RAG produces non-negative accuracy gains across the development
configurations shown in Figure~\ref{fig:phase3_e2e_eval}, although their
magnitude varies substantially across generators, corpus scales, and document
representations. Some MIRIAD configurations approach the dataset-specific Oracle gain, and a
small number exceed it. This is possible because the Oracle condition
uses the evidence supplied by the dataset and is treated as a reference point
rather than as an absolute upper bound.

On MedAESQA, the effect of RAG varies with both the generator and the document
representation. Gemma-3-4B obtains the largest gains overall, including an improvement of
27.50 percentage points in its strongest configuration. Qwen3.5-4B also benefits consistently, whereas MedGemma-1.5-4B
shows greater variation across scales and representations.
Qwen3.5-4B-Think obtains smaller gains overall, alongside its higher LLM-only
accuracy of 85.0\%.

The Abstract representation is generally the strongest and most stable
MedAESQA setting, consistent with the dataset's abstract-level evidence
annotations. TreeRAG is often competitive with or stronger than flat
Full-text chunking. At the \textit{full} PubMed scale, most configurations
still improve over their corresponding LLM-only baselines, although the gains
are generally smaller than at lower corpus scales, particularly for TreeRAG.
This pattern is consistent with the decline in retrieval quality observed as
additional PubMed documents are introduced.

On MIRIAD DEV, RAG yields large gains for Gemma-3-4B and MedGemma-1.5-4B,
whose strongest configurations improve accuracy by 32.00 and 30.50 percentage points, respectively. The two Qwen3.5 configurations also benefit from
retrieval, although their higher LLM-only accuracies leave less room for improvement.

Full-text and TreeRAG generally outperform the Abstract representation on
MIRIAD DEV. Unlike the original passages used to construct Full-text and
TreeRAG units, the Abstract representation consists of synthetically generated
pseudo-abstracts and may omit retrieval-relevant content. TreeRAG remains
competitive with flat Full-text chunking across corpus scales, with only small
differences between them in many configurations. Both representations retain
substantial gains at the \textit{large} MIRIAD scale.

Overall, the development results show that RAG matches or improves upon the
corresponding LLM-only baseline across the evaluated configurations, although
the magnitude of the gains depends on the generator, corpus scale, and document
representation. The next phase evaluates whether these patterns transfer to test datasets that
were not used for pipeline selection.

\subsection{Generalization to Unseen Benchmarks}
\label{sec:results-generalization}

This section evaluates whether the retrieval-augmented pipeline selected during
development transfers beyond the datasets used for pipeline selection. We
consider two test settings. First, five external biomedical MCQA benchmarks
(MMLU-Med, MedQA-US, MedMCQA, PubMedQA, and BioASQ-Y/N) are evaluated using
the \textit{full} PubMed corpus as the retrieval source. Second, the held-out
MIRIAD TEST split is evaluated using the \textit{large} MIRIAD corpus. Although
MIRIAD TEST was not used for pipeline selection, it shares the same construction
process and corpus family as MIRIAD DEV and is therefore interpreted as an
in-domain held-out evaluation rather than as a fully external benchmark.

\begin{table*}[!t]
\centering
\small
\setlength{\tabcolsep}{4pt}
\renewcommand{\arraystretch}{0.95}

\begin{tabular}{@{}lllrrrr@{}}
\toprule
& & & & \multicolumn{3}{c}{\textbf{RAG $\Delta$}} \\
\cmidrule(lr){5-7}
\textbf{Model}
& \textbf{Corpus}
& \textbf{Dataset}
& \textbf{LLM-only}
& Abstract
& Full-text
& TreeRAG \\
\midrule

\multirow[t]{7}{*}{Gemma-3}
& \multirow[t]{6}{*}{PubMed}
& MMLU-Med
& 57.4 & +7.2 & +7.8 & \best{+8.5} \\

&
& MedQA-US
& 46.1 & +11.0 & \best{+13.4} & +10.8 \\

&
& MedMCQA
& 44.9 & +1.8 & +5.0 & \best{+7.4} \\

&
& PubMedQA
& 51.4 & +0.6 & \best{+5.4} & +3.4 \\

&
& BioASQ-Y/N
& 70.2 & \best{+8.3} & +4.7 & \best{+8.3} \\

&
& \cellcolor{gray!15}\textit{Average}
& \cellcolor{gray!15}54.0
& \cellcolor{gray!15}+5.8
& \cellcolor{gray!15}+7.3
& \cellcolor{gray!15}\best{+7.7} \\

\cmidrule(lr){2-7}
& MIRIAD
& MIRIAD TEST
& 65.8 & +21.7 & \best{+30.3} & +29.8 \\

\midrule

\multirow[t]{7}{*}{MedGemma-1.5}
& \multirow[t]{6}{*}{PubMed}
& MMLU-Med
& 53.3 & +6.9 & \best{+7.8} & +6.7 \\

&
& MedQA-US
& 48.3 & +5.2 & \best{+6.0} & +4.6 \\

&
& MedMCQA
& 40.5 & +1.7 & +4.4 & \best{+6.8} \\

&
& PubMedQA
& 41.2 & +9.8 & +10.0 & \best{+15.0} \\

&
& BioASQ-Y/N
& 57.6 & +13.0 & +9.6 & \best{+15.1} \\

&
& \cellcolor{gray!15}\textit{Average}
& \cellcolor{gray!15}48.2
& \cellcolor{gray!15}+7.3
& \cellcolor{gray!15}+7.6
& \cellcolor{gray!15}\best{+9.6} \\

\cmidrule(lr){2-7}
& MIRIAD
& MIRIAD TEST
& 64.0 & +15.9 & +24.8 & \best{+27.4} \\

\midrule

\multirow[t]{7}{*}{Qwen3.5}
& \multirow[t]{6}{*}{PubMed}
& MMLU-Med
& 79.7 & +0.1 & \best{+2.2} & +1.8 \\

&
& MedQA-US
& 70.5 & +1.4 & \best{+4.0} & +1.3 \\

&
& MedMCQA
& 59.4 & \negdelta{$-0.9$} & \best{+3.0} & \best{+3.0} \\

&
& PubMedQA
& 56.0 & \negdelta{$-0.4$} & +1.6 & \best{+2.0} \\

&
& BioASQ-Y/N
& 82.5 & \best{+2.1} & +1.8 & +2.0 \\

&
& \cellcolor{gray!15}\textit{Average}
& \cellcolor{gray!15}69.6
& \cellcolor{gray!15}+0.5
& \cellcolor{gray!15}\best{+2.5}
& \cellcolor{gray!15}+2.0 \\

\cmidrule(lr){2-7}
& MIRIAD
& MIRIAD TEST
& 83.0 & +10.2 & +14.6 & \best{+15.2} \\

\midrule

\multirow[t]{7}{*}{Qwen3.5-Think}
& \multirow[t]{6}{*}{PubMed}
& MMLU-Med
& \best{89.8}
& \negdelta{$-1.7$}
& \negdelta{$-1.6$}
& \negdelta{$-1.3$} \\

&
& MedQA-US
& \best{90.2}
& \negdelta{$-4.0$}
& \negdelta{$-2.9$}
& \negdelta{$-3.6$} \\

&
& MedMCQA
& 69.7
& \negdelta{$-1.1$}
& +0.1
& \best{+0.6} \\

&
& PubMedQA
& \best{56.0}
& \negdelta{$-3.0$}
& \negdelta{$-0.2$}
& \negdelta{$-0.4$} \\

&
& BioASQ-Y/N
& 84.3
& +0.0
& \negdelta{$-1.3$}
& \best{+2.1} \\

&
& \cellcolor{gray!15}\textit{Average}
& \cellcolor{gray!15}\best{78.0}
& \cellcolor{gray!15}\negdelta{$-2.0$}
& \cellcolor{gray!15}\negdelta{$-1.2$}
& \cellcolor{gray!15}\negdelta{$-0.5$} \\

\cmidrule(lr){2-7}
& MIRIAD
& MIRIAD TEST
& 84.1 & +10.5 & +14.4 & \best{+15.0} \\

\bottomrule
\end{tabular}

\caption{End-to-end MCQA performance on external benchmarks and MIRIAD TEST.
LLM-only reports absolute accuracy (\%), while RAG $\Delta$ reports the
percentage-point change relative to the corresponding baseline for each
document representation. PubMed denotes retrieval over the full PubMed corpus,
whereas MIRIAD denotes retrieval over the large MIRIAD corpus. Shaded cells
show the average across the five external benchmarks. Bold marks the largest
improvement in each row, or the LLM-only baseline when no RAG configuration
improves upon it; ties are retained. Negative deltas are shown in red.}
\label{tab:test_e2e_delta}
\end{table*}

On the five external benchmarks, Gemma-3 and MedGemma-1.5 improve over their
LLM-only baselines across every dataset and evaluated representation. Their
largest average gains are obtained with TreeRAG, reaching +7.7 and +9.6
percentage points, respectively. Qwen3.5 also benefits from RAG, although its
average gains are smaller: Full-text performs best at +2.5 points, followed by
TreeRAG at +2.0 points. For Qwen3.5-Think, RAG produces negative average gains across all three representations, although TreeRAG performs best, with an average change of $-0.5$ points, compared with $-1.2$ for Full-text and $-2.0$ for Abstract. TreeRAG improves performance on MedMCQA and BioASQ-Y/N, but these gains do not fully offset the decreases observed on MMLU-Med, MedQA-US, and PubMedQA.

The representation trends differ from those observed on MedAESQA DEV. Abstract
retrieval was generally strongest during development, where the reference
evidence is abstract-based, whereas Full-text and TreeRAG obtain the strongest
external results for most generators and benchmarks. TreeRAG achieves the best
average gain for Gemma-3 and MedGemma-1.5 and remains competitive with
Full-text for Qwen3.5, despite using fewer indexed retrieval units than flat
full-text chunking. BioASQ-Y/N is a notable exception to the broader pattern:
Abstract remains competitive, while TreeRAG matches or exceeds it for some
generators. Overall, these results indicate that representation performance
depends on both the benchmark and the generator rather than on a single
representation being consistently optimal.

The held-out MIRIAD TEST results preserve the main patterns observed on MIRIAD
DEV. RAG improves all four generators, with gains ranging from +10.2 to +30.3
percentage points. Abstract produces the smallest gain for every generator,
whereas Full-text and TreeRAG are consistently stronger. TreeRAG obtains the
largest gain for three of the four generators and remains within 0.5 points of
Full-text for Gemma-3. This pattern is consistent with the construction of the
MIRIAD representations: the Abstract condition uses synthetically generated
pseudo-abstracts, whereas Full-text and TreeRAG preserve the original passage
content more directly.

Overall, the selected RAG pipeline transfers successfully to the external
benchmarks for Gemma-3, MedGemma-1.5, and Qwen3.5, although the magnitude of
the gains and the strongest representation vary across generators and
datasets. TreeRAG is particularly notable, achieving the strongest average
external gains for Gemma-3 and MedGemma-1.5 and the largest MIRIAD TEST gain
for three of the four generators, while remaining competitive with Full-text
elsewhere. The development-stage behavior also transfers clearly to the
in-domain MIRIAD TEST split, where all generators improve and the
full-text-derived representations remain substantially stronger than the
pseudo-abstract representation.

\subsection{Retrieval Efficiency and Trade-offs}
\label{sec:results-efficiency}

Beyond retrieval quality and downstream accuracy, large-scale biomedical RAG
systems are constrained by query-time latency and index storage. We analyze
these trade-offs on MedAESQA using the \textit{full} PubMed corpus, the largest
biomedical retrieval setting evaluated in this study.

Latency is evaluated on MedAESQA using the \textit{full} PubMed Abstract
index, the strongest representation in the MedAESQA development retrieval
experiments.

\begin{figure*}[!t]
\centering
\includegraphics[width=\textwidth]
{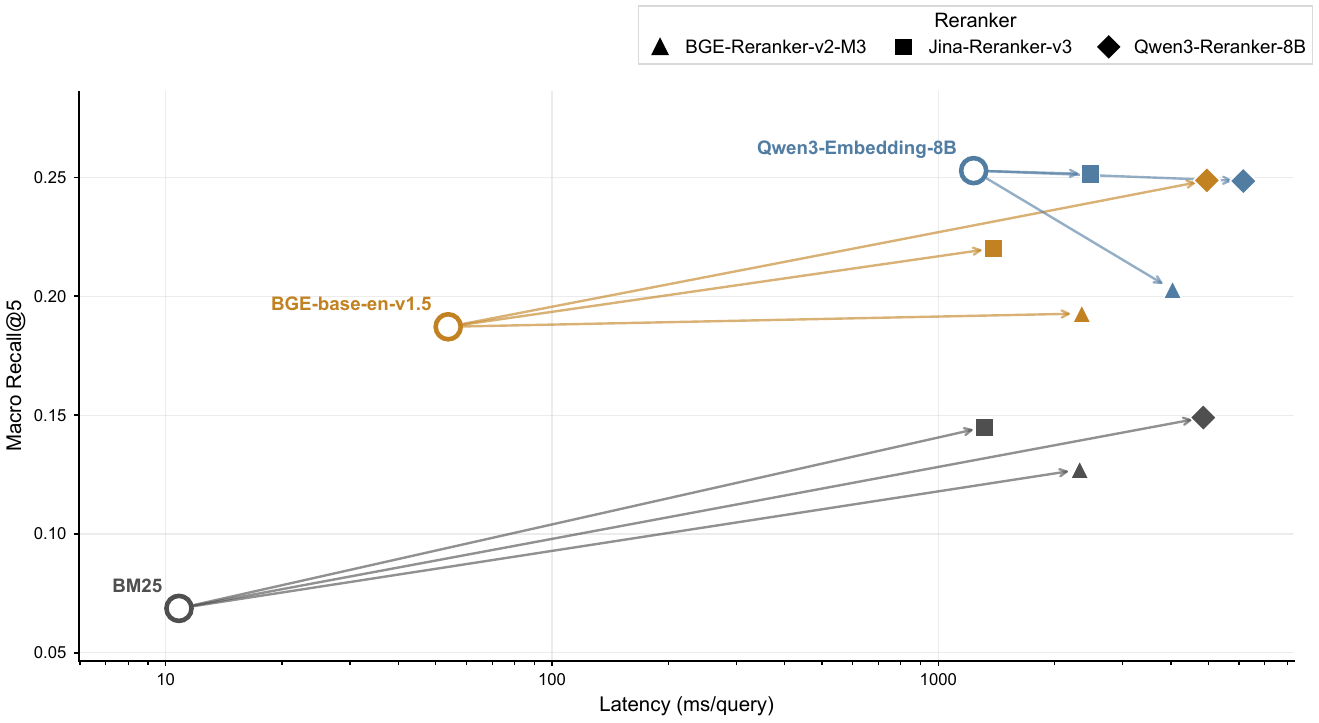}
\caption{Efficiency--effectiveness trade-off on MedAESQA using the
\textit{full} PubMed Abstract index. The x-axis reports mean per-query latency
on a logarithmic scale, and the y-axis reports Macro Recall@5. Open circles
denote first-stage retrievers, while filled markers denote the corresponding
configurations after reranking the top 100 candidates. Colors identify the
first-stage retriever, marker shapes identify the reranker, and arrows connect
each retriever to its reranked variants.}
\label{fig:medaesqa_latency_recall5}
\end{figure*}

Figure~\ref{fig:medaesqa_latency_recall5} reports Macro Recall@5 against mean
per-query latency for the first-stage and reranked configurations.

The first-stage retrievers exhibit a clear latency--effectiveness trade-off.
BM25 has the lowest mean latency at 10.8~ms per query but obtains substantially
lower Recall@5 than the dense retrievers. BGE-base-en-v1.5 increases latency to
53.9~ms while providing a marked improvement in retrieval effectiveness.
Qwen3-Embedding-8B achieves the highest first-stage Macro Recall@5, reaching
25.28\%, but requires 1234.7~ms per query. Thus, BGE-base-en-v1.5 is
approximately 23 times faster than Qwen3-Embedding-8B, although it does not
reach the same retrieval effectiveness.

Reranking introduces a substantial latency overhead for every first-stage
retriever. For BM25, reranking improves Recall@5 but moves the resulting
pipelines into the latency range of substantially more effective dense
retrieval configurations. Reranking BGE-base-en-v1.5 can further improve its
Recall@5, but only at a query-time cost that exceeds that of
Qwen3-Embedding-8B without reranking. For Qwen3-Embedding-8B, none of the
rerankers improves upon the first-stage Recall@5, and some reduce it while
increasing latency. Consequently, the three first-stage retrievers define the
observed latency--effectiveness frontier: Qwen3-Embedding-8B provides maximum
effectiveness, BGE-base-en-v1.5 offers a substantially lower-latency
alternative, and BM25 minimizes latency at the cost of lower recall.

Storage requirements also vary substantially across representations and
retrievers. At the \textit{full} PubMed scale, Abstract contains 4.3M indexed
retrieval units and requires 5.9, 18.8, and 73.3~GB for BM25,
BGE-base-en-v1.5, and Qwen3-Embedding-8B, respectively. Full-text chunking
expands the collection to 117.4M units, increasing the corresponding index
sizes to 199.5, 481.0, and 1985.1~GB. TreeRAG indexes 38.3M units and requires
111.9, 228.1, and 872.0~GB. Relative to flat Full-text chunking, TreeRAG uses
approximately 67\% fewer retrieval units and reduces storage by 44--56\%,
depending on the retriever.

Overall, maximum retrieval effectiveness is obtained with
Qwen3-Embedding-8B without reranking, whereas BGE-base-en-v1.5 provides a
substantially lower-latency alternative. TreeRAG offers a complementary
storage trade-off: it requires considerably fewer indexed retrieval units than
Full-text while remaining competitive with, and often outperforming, flat
Full-text chunking in the downstream evaluations.


\section{Discussion}
\label{sec:discussion}

Our results show that biomedical RAG design choices should be evaluated in
relation to corpus scale. Larger retrieval pools broaden the available evidence
but also introduce additional distractors, making evidence selection more
difficult. The degradation observed for several retrieval configurations at the
largest corpus scales shows that conclusions obtained from small or controlled
collections do not necessarily transfer to large-scale biomedical retrieval.
Corpus scale should therefore be treated as an explicit evaluation variable and
considered jointly with document representation, retrieval quality, generator
behavior, and deployment cost.

Document representation strongly shapes this interaction. Abstract retrieval
performs particularly well on MedAESQA, whose reference evidence is provided at
the abstract level. In contrast, Full-text and TreeRAG generally produce
stronger gains on the external benchmarks evaluated using the \textit{full}
PubMed corpus. MIRIAD shows an even clearer preference for representations that
preserve the original passage content: its synthetically generated
pseudo-abstracts are consistently weaker than Full-text and TreeRAG. These
contrasting patterns indicate that no document representation is universally
optimal. Its effectiveness depends on the evidence requirements of the
benchmark and on how the retrieved units interact with the generator.

TreeRAG provides a particularly relevant effectiveness--efficiency trade-off.
It achieves the strongest average external gains for Gemma-3 and
MedGemma-1.5, remains competitive with Full-text for Qwen3.5, and obtains the
largest MIRIAD TEST gain for three of the four generators. At the same time,
TreeRAG indexes 38.3M retrieval units, compared with 117.4M for flat Full-text
chunking, reducing storage requirements by 44--56\% depending on the
retriever. These findings show that competitive retrieval over full-text content
does not necessarily require indexing the complete set of fixed-length chunks.
However, the relative advantage of TreeRAG still varies across generators,
benchmarks, and corpus scales, and the present experiments do not isolate
hierarchical organization as the sole cause of its performance.

For Qwen3.5-Think, TreeRAG produces the strongest external average among the
three representations, although its mean change remains slightly negative at
$-0.5$ percentage points. Its gains on MedMCQA and BioASQ-Y/N are offset by
decreases on MMLU-Med, MedQA-US, and PubMedQA. This mixed behavior further
indicates that the effectiveness of TreeRAG depends on the interaction between
the generator and the benchmark.

The results also challenge the assumption that reranking is an inherently
beneficial stage. On MedAESQA, Qwen3-Embedding-8B obtains the strongest
Recall@5 without reranking, and the evaluated rerankers either preserve or
reduce effectiveness while increasing latency. On MIRIAD, reranking becomes
more useful as the retrieval pool grows, and Qwen3-Reranker-8B improves the
final evidence ranking at the largest scale. The utility of reranking therefore
depends on the quality of the first-stage ranking, the corpus, and the evidence
units being ranked. Its value should be established empirically rather than
assumed from improvements observed in smaller or different retrieval settings.

Downstream performance further depends on the generator. Gemma-3 and
MedGemma-1.5 obtain the largest RAG gains, whereas Qwen3.5 shows smaller but
mostly positive improvements on the external benchmarks. Qwen3.5-Think benefits
strongly from RAG on MIRIAD TEST but shows smaller, mixed, or negative changes
on the external benchmarks evaluated with PubMed retrieval. A strong LLM-only
baseline reduces the available margin for improvement, but baseline performance
alone does not explain whether retrieval will help: the same generator can
respond differently across benchmarks and document representations. Retrieval
and generation should therefore be evaluated jointly rather than treating
retrieval metrics as sufficient predictors of end-to-end performance.

Finally, the efficiency analysis illustrates that the most effective
configuration is not necessarily the most practical one. Qwen3-Embedding-8B
provides the highest first-stage retrieval effectiveness but has substantially
higher latency than BGE-base-en-v1.5, while reranking introduces additional
cost without consistently improving Recall@5. Similarly, Full-text supports
strong downstream results but produces the largest indexes, whereas TreeRAG
retains competitive performance with a considerably smaller storage footprint.
The preferred biomedical RAG configuration therefore depends on the intended
deployment setting: maximum-effectiveness systems may justify larger models and
indexes, whereas latency- or storage-constrained applications may favor lighter
retrievers or more compact full-text representations.

\subsection{Limitations}
\label{sec:limitations}

This study has some limitations related to dataset scope and evaluation
design. MedAESQA contains relatively few questions, whereas MIRIAD provides a
larger but synthetically constructed setting. Moreover, MIRIAD TEST shares the
same construction process and corpus family as MIRIAD DEV and should therefore
be interpreted as an in-domain held-out evaluation. We complement these
controlled datasets with five independently constructed biomedical benchmarks
to assess transfer beyond the development settings.

The evaluation is restricted to multiple-choice question answering, enabling
standardized comparisons across corpus scales, retrieval configurations, and
generators. It does not directly assess open-ended generation, explanation
quality, evidence citation, or clinical validity. In addition, retrieval is
evaluated through source-document matching because the gold evidence formats
differ across datasets and do not align exactly with all retrieval units.
Recall should therefore be interpreted as source-level evidence coverage
rather than exact evidence-unit matching.

Finally, the reported latency and storage values depend on the hardware,
software, and indexing configuration used in this study. They support
controlled relative comparisons between configurations but should not be
interpreted as universal deployment costs.

\section{Conclusion}
\label{sec:conclusion}

This work addresses an important limitation of existing biomedical RAG
evaluation: systems are commonly assessed at a single fixed corpus scale, even
when the retrieval index itself is large. Such evaluations cannot establish
whether an observed advantage remains robust as the collection grows. Under
controlled scaling across PubMed and MIRIAD, from the smallest conditions to
the full PubMed corpus and the complete MIRIAD collection, advantages observed
for retrievers, rerankers, and document representations at the control or tiny
scales did not consistently persist at larger scales, despite retaining the
available gold sources. Corpus scale should therefore be treated as an explicit
evaluation variable rather than assuming that results obtained at one
collection size will generalize to another.

No document representation was consistently optimal. Abstract retrieval was
effective when the reference evidence was abstract-aligned, whereas Full-text
and TreeRAG were generally stronger on the external benchmarks and MIRIAD.
TreeRAG provided a favorable effectiveness--efficiency trade-off, remaining
competitive with, and frequently outperforming, flat Full-text chunking while
using approximately 67\% fewer indexed retrieval units and 44--56\% less
storage. Reranking was likewise dataset-dependent, improving evidence selection
at the largest MIRIAD scale but providing no effectiveness advantage for MedAESQA retrieval over the full PubMed corpus.

Overall, this study provides a controlled, large-scale empirical
characterization of biomedical RAG under corpus growth and operational
constraints. By jointly evaluating retrieval, reranking, generation, latency,
and storage, the results provide evidence for selecting RAG configurations
according to effectiveness and computational cost.


\FloatBarrier

\section*{CRediT Authorship Contribution Statement}

\textbf{Adrián Sánchez-Freire}: Conceptualization, Methodology,
Software, Investigation, Formal analysis, Validation,
Visualization, Writing -- original draft, Writing -- review
and editing. \textbf{Gonzalo Pérez-Gómez}: Conceptualization, Methodology,
Software, Investigation, Writing -- review and editing. \textbf{Aitziber Atutxa}: Conceptualization, Methodology, Supervision,
Writing -- review and editing. \textbf{Ander Barrena}: Conceptualization, Methodology, Supervision,
Writing -- review and editing.

\section*{Ethics statement}
This study used biomedical literature and previously released benchmark
datasets. It did not involve the recruitment of human participants, access to
identifiable personal data, or animal experimentation. Therefore, ethics
committee approval and informed consent were not required.

\section*{Data Availability}
The source code, experimental scripts, configurations, and resources supporting
this study are publicly available in two GitHub repositories. The main
experimental repository is available at
\url{https://github.com/hitz-zentroa/biomedical-rag-scaling}. The pipeline used to
convert open-ended question-answering datasets into multiple-choice question
answering format is available at
\url{https://github.com/hitz-zentroa/qa-to-mcqa}.

\section*{Acknowledgements}
This work is supported by \seqsplit{MCIN/AEI/10.13039/501100011033} and FEDER/UE
via EDHER-MED--EDHIA (PID2022-136522OB-C22),
TRUST-MED (PID2025-174880OB-I00),
and DeepKnowledge (PID2021-127777OB-C21).
Secondary support is provided by the HiTZ Center
and the Basque Government (IT1570-22),
and EFA 104/01-LINGUATEC IA (ERDF/INTERREG POCTEFA 2021--2027).
Adrián Sánchez-Freire is supported by a predoctoral research contract
from the University of the Basque Country (EHU), reference PIF25/58. The majority of the inference experiments were carried out on the Leonardo supercomputer at CINECA through resources provided by the EuroHPC Joint Undertaking, project EHPC-AIF2026SC01-156. 

The funders had no role in the study design; data collection,
analysis, or interpretation; preparation of the manuscript;
or the decision to submit the article for publication.

\section*{Declaration of Generative AI and AI-Assisted Technologies in the Manuscript Preparation Process}
During the preparation of this work, the authors used
ChatGPT (OpenAI) for language editing. After using this
tool, the authors reviewed and edited the content as needed
and take full responsibility for the content of the
published article.

\bibliographystyle{elsarticle-num}
\bibliography{cas-refs}

@inproceedings{lewis2021rag,
 author = {Lewis, Patrick and Perez, Ethan and Piktus, Aleksandra and Petroni, Fabio and Karpukhin, Vladimir and Goyal, Naman and K\"{u}ttler, Heinrich and Lewis, Mike and Yih, Wen-tau and Rockt\"{a}schel, Tim and Riedel, Sebastian and Kiela, Douwe},
 booktitle = {Advances in Neural Information Processing Systems},
 editor = {H. Larochelle and M. Ranzato and R. Hadsell and M.F. Balcan and H. Lin},
 pages = {9459--9474},
 publisher = {Curran Associates, Inc.},
 title = {Retrieval-Augmented Generation for Knowledge-Intensive NLP Tasks},
 url = {https://proceedings.neurips.cc/paper_files/paper/2020/file/6b493230205f780e1bc26945df7481e5-Paper.pdf},
 volume = {33},
 year = {2020}
}

@misc{gao2023rag_survey,
      title={Retrieval-Augmented Generation for Large Language Models: A Survey}, 
      author={Yunfan Gao and Yun Xiong and Xinyu Gao and Kangxiang Jia and Jinliu Pan and Yuxi Bi and Yi Dai and Jiawei Sun and Meng Wang and Haofen Wang},
      year={2024},
      eprint={2312.10997},
      archivePrefix={arXiv},
      primaryClass={cs.CL},
      url={https://arxiv.org/abs/2312.10997}, 
}

@article{zhao2024rag_survey,
  title = {Retrieval-Augmented Generation for AI-Generated Content: A Survey},
  author = {Zhao, Penghao and Zhang, Hailin and Yu, Qinhan and Wang, Zhengren and Geng, Yunteng and Fu, Fangcheng and Yang, Ling and Zhang, Wentao and Jiang, Jie and Cui, Bin},
  year = {2026},
  journal = {Data Science and Engineering},
  volume = {11},
  number = {1},
  pages = {1--29},
  doi = {10.1007/s41019-025-00335-5},
  url = {https://doi.org/10.1007/s41019-025-00335-5},
  isbn = {2364-1541},
}

@inproceedings{vladika2024healthqa,
  title     = {Improving Health Question Answering with Reliable and Time-Aware Evidence Retrieval},
  author    = {Vladika, Juraj and Matthes, Florian},
  booktitle = {Findings of the Association for Computational Linguistics: NAACL 2024},
  pages     = {4752--4763},
  year      = {2024},
  address   = {Mexico City, Mexico},
  publisher = {Association for Computational Linguistics},
  doi       = {10.18653/v1/2024.findings-naacl.295},
  url       = {https://aclanthology.org/2024.findings-naacl.295/}
}

@article{li2025biomedrag,
title = {BiomedRAG: A retrieval augmented large language model for biomedicine},
journal = {Journal of Biomedical Informatics},
volume = {162},
pages = {104769},
year = {2025},
issn = {1532-0464},
doi = {https://doi.org/10.1016/j.jbi.2024.104769},
url = {https://www.sciencedirect.com/science/article/pii/S1532046424001874},
author = {Mingchen Li and Halil Kilicoglu and Hua Xu and Rui Zhang},
}

@article{lopez2025clear,
  title   = {Clinical Entity Augmented Retrieval for Clinical Information Extraction},
  author  = {Lopez, Ivan and Swaminathan, Akshay and Vedula, Karthik and Narayanan, Sanjana and Nateghi Haredasht, Fateme and Ma, Stephen P. and Liang, April S. and Tate, Steven and Maddali, Manoj and Gallo, Robert Joseph and Shah, Nigam H. and Chen, Jonathan H.},
  journal = {npj Digital Medicine},
  volume  = {8},
  number  = {1},
  pages   = {45},
  year    = {2025},
  doi     = {10.1038/s41746-024-01377-1},
  url     = {https://doi.org/10.1038/s41746-024-01377-1}
}

@inproceedings{sohn2025rag2,
    title = "Rationale-Guided Retrieval Augmented Generation for Medical Question Answering",
    author = "Sohn, Jiwoong  and
      Park, Yein  and
      Yoon, Chanwoong  and
      Park, Sihyeon  and
      Hwang, Hyeon  and
      Sung, Mujeen  and
      Kim, Hyunjae  and
      Kang, Jaewoo",
    editor = "Chiruzzo, Luis  and
      Ritter, Alan  and
      Wang, Lu",
    booktitle = "Proceedings of the 2025 Conference of the Nations of the Americas Chapter of the Association for Computational Linguistics: Human Language Technologies (Volume 1: Long Papers)",
    month = apr,
    year = "2025",
    address = "Albuquerque, New Mexico",
    publisher = "Association for Computational Linguistics",
    url = "https://aclanthology.org/2025.naacl-long.635/",
    doi = "10.18653/v1/2025.naacl-long.635",
    pages = "12739--12753",
    ISBN = "979-8-89176-189-6",
}

@InProceedings{borgeaud2022retro,
  title = 	 {Improving Language Models by Retrieving from Trillions of Tokens},
  author =       {Borgeaud, Sebastian and Mensch, Arthur and Hoffmann, Jordan and Cai, Trevor and Rutherford, Eliza and Millican, Katie and Van Den Driessche, George Bm and Lespiau, Jean-Baptiste and Damoc, Bogdan and Clark, Aidan and De Las Casas, Diego and Guy, Aurelia and Menick, Jacob and Ring, Roman and Hennigan, Tom and Huang, Saffron and Maggiore, Loren and Jones, Chris and Cassirer, Albin and Brock, Andy and Paganini, Michela and Irving, Geoffrey and Vinyals, Oriol and Osindero, Simon and Simonyan, Karen and Rae, Jack and Elsen, Erich and Sifre, Laurent},
  booktitle = 	 {Proceedings of the 39th International Conference on Machine Learning},
  pages = 	 {2206--2240},
  year = 	 {2022},
  editor = 	 {Chaudhuri, Kamalika and Jegelka, Stefanie and Song, Le and Szepesvari, Csaba and Niu, Gang and Sabato, Sivan},
  volume = 	 {162},
  series = 	 {Proceedings of Machine Learning Research},
  month = 	 {17--23 Jul},
  publisher =    {PMLR},
  url = 	 {https://proceedings.mlr.press/v162/borgeaud22a.html},
}

@article{ram2023context,
    title = "In-Context Retrieval-Augmented Language Models",
    author = "Ram, Ori  and
      Levine, Yoav  and
      Dalmedigos, Itay  and
      Muhlgay, Dor  and
      Shashua, Amnon  and
      Leyton-Brown, Kevin  and
      Shoham, Yoav",
    journal = "Transactions of the Association for Computational Linguistics",
    volume = "11",
    year = "2023",
    address = "Cambridge, MA",
    publisher = "MIT Press",
    url = "https://aclanthology.org/2023.tacl-1.75/",
    doi = "10.1162/tacl_a_00605",
    pages = "1316--1331",
}

@inproceedings{jiang2023active,
    title = "Active Retrieval Augmented Generation",
    author = "Jiang, Zhengbao  and
      Xu, Frank  and
      Gao, Luyu  and
      Sun, Zhiqing  and
      Liu, Qian  and
      Dwivedi-Yu, Jane  and
      Yang, Yiming  and
      Callan, Jamie  and
      Neubig, Graham",
    editor = "Bouamor, Houda  and
      Pino, Juan  and
      Bali, Kalika",
    booktitle = "Proceedings of the 2023 Conference on Empirical Methods in Natural Language Processing",
    month = dec,
    year = "2023",
    address = "Singapore",
    publisher = "Association for Computational Linguistics",
    url = "https://aclanthology.org/2023.emnlp-main.495/",
    doi = "10.18653/v1/2023.emnlp-main.495",
    pages = "7969--7992",
}

@article{
mialon2023augmented,
title={Augmented Language Models: a Survey},
author={Gr{\'e}goire Mialon and Roberto Dessi and Maria Lomeli and Christoforos Nalmpantis and Ramakanth Pasunuru and Roberta Raileanu and Baptiste Roziere and Timo Schick and Jane Dwivedi-Yu and Asli Celikyilmaz and Edouard Grave and Yann LeCun and Thomas Scialom},
journal={Transactions on Machine Learning Research},
issn={2835-8856},
year={2023},
url={https://openreview.net/forum?id=jh7wH2AzKK},
note={Survey Certification}
}

@article{Jacquemart2003medicalqa,
  title    = "Towards a medical question-answering system: a feasibility study",
  author   = "Jacquemart, Pierre and Zweigenbaum, Pierre",
  journal  = "Stud. Health Technol. Inform.",
  volume   =  95,
  pages    = "463--468",
  year     =  2003,
  language = "en"
}

@article{lee2020biobert,
   title={BioBERT: a pre-trained biomedical language representation model for biomedical text mining},
   volume={36},
   ISSN={1367-4811},
   url={http://dx.doi.org/10.1093/bioinformatics/btz682},
   DOI={10.1093/bioinformatics/btz682},
   number={4},
   journal={Bioinformatics},
   publisher={Oxford University Press (OUP)},
   author={Lee, Jinhyuk and Yoon, Wonjin and Kim, Sungdong and Kim, Donghyeon and Kim, Sunkyu and So, Chan Ho and Kang, Jaewoo},
   editor={Wren, Jonathan},
   year={2019},
   month={Sept}, pages={1234–1240} }

@article{otegi2022covidqa,
title = {Information retrieval and question answering: A case study on COVID-19 scientific literature},
journal = {Knowledge-Based Systems},
volume = {240},
pages = {108072},
year = {2022},
issn = {0950-7051},
doi = {https://doi.org/10.1016/j.knosys.2021.108072},
url = {https://www.sciencedirect.com/science/article/pii/S0950705121011485},
author = {Arantxa Otegi and Iñaki {San Vicente} and Xabier Saralegi and Anselmo Peñas and Borja Lozano and Eneko Agirre}
}

@article{medexpqa,
title = {MedExpQA: Multilingual benchmarking of Large Language Models for Medical Question Answering},
journal = {Artificial Intelligence in Medicine},
volume = {155},
pages = {102938},
year = {2024},
issn = {0933-3657},
doi = {https://doi.org/10.1016/j.artmed.2024.102938},
url = {https://www.sciencedirect.com/science/article/pii/S0933365724001805},
author = {Iñigo Alonso and Maite Oronoz and Rodrigo Agerri},
}

@article{athenikos2010,
title = {Biomedical question answering: A survey},
journal = {Computer Methods and Programs in Biomedicine},
volume = {99},
number = {1},
pages = {1-24},
year = {2010},
issn = {0169-2607},
doi = {https://doi.org/10.1016/j.cmpb.2009.10.003},
url = {https://www.sciencedirect.com/science/article/pii/S0169260709002879},
author = {Sofia J. Athenikos and Hyoil Han}}

@article{jin2022survey,
   title={Biomedical Question Answering: A Survey of Approaches and Challenges},
   volume={55},
   ISSN={1557-7341},
   url={http://dx.doi.org/10.1145/3490238},
   DOI={10.1145/3490238},
   number={2},
   journal={ACM Computing Surveys},
   publisher={Association for Computing Machinery (ACM)},
   author={Jin, Qiao and Yuan, Zheng and Xiong, Guangzhi and Yu, Qianlan and Ying, Huaiyuan and Tan, Chuanqi and Chen, Mosha and Huang, Songfang and Liu, Xiaozhong and Yu, Sheng},
   year={2022},
   month=Jan, pages={1–36} }

@article{gu2021domain,
   title={Domain-Specific Language Model Pretraining for Biomedical Natural Language Processing},
   volume={3},
   ISSN={2637-8051},
   url={http://dx.doi.org/10.1145/3458754},
   DOI={10.1145/3458754},
   number={1},
   journal={ACM Transactions on Computing for Healthcare},
   publisher={Association for Computing Machinery (ACM)},
   author={Gu, Yu and Tinn, Robert and Cheng, Hao and Lucas, Michael and Usuyama, Naoto and Liu, Xiaodong and Naumann, Tristan and Gao, Jianfeng and Poon, Hoifung},
   year={2021},
   month=Oct, pages={1–23} }

@inproceedings{yasunaga2022linkbert,
    title = "{L}ink{BERT}: Pretraining Language Models with Document Links",
    author = "Yasunaga, Michihiro  and
      Leskovec, Jure  and
      Liang, Percy",
    editor = "Muresan, Smaranda  and
      Nakov, Preslav  and
      Villavicencio, Aline",
    booktitle = "Proceedings of the 60th Annual Meeting of the Association for Computational Linguistics (Volume 1: Long Papers)",
    month = may,
    year = "2022",
    address = "Dublin, Ireland",
    publisher = "Association for Computational Linguistics",
    url = "https://aclanthology.org/2022.acl-long.551/",
    doi = "10.18653/v1/2022.acl-long.551",
    pages = "8003--8016",
}

@article{singhal2023medpalm,
  title = {Large language models encode clinical knowledge},
  author = {Singhal, Karan and Azizi, Shekoofeh and Tu, Tao and Mahdavi, S. Sara and Wei, Jason and Chung, Hyung Won and Scales, Nathan and Tanwani, Ajay and Cole-Lewis, Heather and Pfohl, Stephen and Payne, Perry and Seneviratne, Martin and Gamble, Paul and Kelly, Chris and Babiker, Abubakr and Schärli, Nathanael and Chowdhery, Aakanksha and Mansfield, Philip and Demner-Fushman, Dina and Agüera y Arcas, Blaise and Webster, Dale and Corrado, Greg S. and Matias, Yossi and Chou, Katherine and Gottweis, Juraj and Tomasev, Nenad and Liu, Yun and Rajkomar, Alvin and Barral, Joelle and Semturs, Christopher and Karthikesalingam, Alan and Natarajan, Vivek},
  year = {2023},
  journal = {Nature},
  volume = {620},
  number = {7972},
  pages = {172--180},
  doi = {10.1038/s41586-023-06291-2},
  url = {https://doi.org/10.1038/s41586-023-06291-2},
  isbn = {1476-4687},
}

@article{singhal2023medpalm2,
  title = {Toward expert-level medical question answering with large language models},
  author = {Singhal, Karan and Tu, Tao and Gottweis, Juraj and Sayres, Rory and Wulczyn, Ellery and Amin, Mohamed and Hou, Le and Clark, Kevin and Pfohl, Stephen R. and Cole-Lewis, Heather and Neal, Darlene and Rashid, Qazi Mamunur and Schaekermann, Mike and Wang, Amy and Dash, Dev and Chen, Jonathan H. and Shah, Nigam H. and Lachgar, Sami and Mansfield, Philip Andrew and Prakash, Sushant and Green, Bradley and Dominowska, Ewa and Agüera y Arcas, Blaise and Tomašev, Nenad and Liu, Yun and Wong, Renee and Semturs, Christopher and Mahdavi, S. Sara and Barral, Joelle K. and Webster, Dale R. and Corrado, Greg S. and Matias, Yossi and Azizi, Shekoofeh and Karthikesalingam, Alan and Natarajan, Vivek},
  year = {2025},
  journal = {Nature Medicine},
  volume = {31},
  number = {3},
  pages = {943--950},
  doi = {10.1038/s41591-024-03423-7},
  url = {https://doi.org/10.1038/s41591-024-03423-7},
  isbn = {1546-170X},
}

@misc{nori2023capabilities,
      title={Capabilities of GPT-4 on Medical Challenge Problems}, 
      author={Harsha Nori and Nicholas King and Scott Mayer McKinney and Dean Carignan and Eric Horvitz},
      year={2023},
      eprint={2303.13375},
      archivePrefix={arXiv},
      primaryClass={cs.CL},
      url={https://arxiv.org/abs/2303.13375}, 
}

@inproceedings{karpukhin2020dpr,
    title = "Dense Passage Retrieval for Open-Domain Question Answering",
    author = "Karpukhin, Vladimir  and
      Oguz, Barlas  and
      Min, Sewon  and
      Lewis, Patrick  and
      Wu, Ledell  and
      Edunov, Sergey  and
      Chen, Danqi  and
      Yih, Wen-tau",
    editor = "Webber, Bonnie  and
      Cohn, Trevor  and
      He, Yulan  and
      Liu, Yang",
    booktitle = "Proceedings of the 2020 Conference on Empirical Methods in Natural Language Processing (EMNLP)",
    month = nov,
    year = "2020",
    address = "Online",
    publisher = "Association for Computational Linguistics",
    url = "https://aclanthology.org/2020.emnlp-main.550/",
    doi = "10.18653/v1/2020.emnlp-main.550",
    pages = "6769--6781",
}

@inproceedings{izacard2021fid,
    title = "Leveraging Passage Retrieval with Generative Models for Open Domain Question Answering",
    author = "Izacard, Gautier  and
      Grave, Edouard",
    editor = "Merlo, Paola  and
      Tiedemann, Jorg  and
      Tsarfaty, Reut",
    booktitle = "Proceedings of the 16th Conference of the European Chapter of the Association for Computational Linguistics: Main Volume",
    month = apr,
    year = "2021",
    address = "Online",
    publisher = "Association for Computational Linguistics",
    url = "https://aclanthology.org/2021.eacl-main.74/",
    doi = "10.18653/v1/2021.eacl-main.74",
    pages = "874--880",
}

@inproceedings{liu2021dense_hierarchical,
    title = "Dense Hierarchical Retrieval for Open-domain Question Answering",
    author = "Liu, Ye  and
      Hashimoto, Kazuma  and
      Zhou, Yingbo  and
      Yavuz, Semih  and
      Xiong, Caiming  and
      Yu, Philip",
    editor = "Moens, Marie-Francine  and
      Huang, Xuanjing  and
      Specia, Lucia  and
      Yih, Scott Wen-tau",
    booktitle = "Findings of the Association for Computational Linguistics: EMNLP 2021",
    month = nov,
    year = "2021",
    address = "Punta Cana, Dominican Republic",
    publisher = "Association for Computational Linguistics",
    url = "https://aclanthology.org/2021.findings-emnlp.19/",
    doi = "10.18653/v1/2021.findings-emnlp.19",
    pages = "188--200",
}

@misc{bhat2025rethinking_chunk_size,
      title={Rethinking Chunk Size For Long-Document Retrieval: A Multi-Dataset Analysis}, 
      author={Sinchana Ramakanth Bhat and Max Rudat and Jannis Spiekermann and Nicolas Flores-Herr},
      year={2025},
      eprint={2505.21700},
      archivePrefix={arXiv},
      primaryClass={cs.IR},
      url={https://arxiv.org/abs/2505.21700}, 
}

@inproceedings{dasigi2021qasper,
    title = "A Dataset of Information-Seeking Questions and Answers Anchored in Research Papers",
    author = "Dasigi, Pradeep  and
      Lo, Kyle  and
      Beltagy, Iz  and
      Cohan, Arman  and
      Smith, Noah A.  and
      Gardner, Matt",
    editor = "Toutanova, Kristina  and
      Rumshisky, Anna  and
      Zettlemoyer, Luke  and
      Hakkani-Tur, Dilek  and
      Beltagy, Iz  and
      Bethard, Steven  and
      Cotterell, Ryan  and
      Chakraborty, Tanmoy  and
      Zhou, Yichao",
    booktitle = "Proceedings of the 2021 Conference of the North American Chapter of the Association for Computational Linguistics: Human Language Technologies",
    month = jun,
    year = "2021",
    address = "Online",
    publisher = "Association for Computational Linguistics",
    url = "https://aclanthology.org/2021.naacl-main.365/",
    doi = "10.18653/v1/2021.naacl-main.365",
    pages = "4599--4610"
}

@inproceedings{
sarthi2024raptor,
title={{RAPTOR}: Recursive Abstractive Processing for Tree-Organized Retrieval},
author={Parth Sarthi and Salman Abdullah and Aditi Tuli and Shubh Khanna and Anna Goldie and Christopher D Manning},
booktitle={The Twelfth International Conference on Learning Representations},
year={2024},
url={https://openreview.net/forum?id=GN921JHCRw}
}

@misc{jiang2024longrag,
      title={LongRAG: Enhancing Retrieval-Augmented Generation with Long-context LLMs}, 
      author={Ziyan Jiang and Xueguang Ma and Wenhu Chen},
      year={2024},
      eprint={2406.15319},
      archivePrefix={arXiv},
      primaryClass={cs.CL},
      url={https://arxiv.org/abs/2406.15319}, 
}

@article{izacard2023atlas,
  author  = {Gautier Izacard and Patrick Lewis and Maria Lomeli and Lucas Hosseini and Fabio Petroni and Timo Schick and Jane Dwivedi-Yu and Armand Joulin and Sebastian Riedel and Edouard Grave},
  title   = {Atlas: Few-shot Learning with Retrieval Augmented Language Models},
  journal = {Journal of Machine Learning Research},
  year    = {2023},
  volume  = {24},
  number  = {251},
  pages   = {1--43},
  url     = {http://jmlr.org/papers/v24/23-0037.html}
}

@InProceedings{less_llm_more_documents,
author="Ning, Jingjie
and Kong, Yibo
and Long, Yunfan
and Callan, Jamie",
editor="Campos, Ricardo
and Jatowt, Adam
and Lan, Yanyan
and Aliannejadi, Mohammad
and Bauer, Christine
and MacAvaney, Sean
and Anand, Avishek
and Ren, Zhaochun
and Verberne, Suzan
and Bai, Nan
and Mansoury, Masoud",
title="Less LLM, More Documents: Searching for Improved RAG",
booktitle="Advances in Information Retrieval",
year="2026",
publisher="Springer Nature Switzerland",
address="Cham",
pages="598--613",
isbn="978-3-032-21289-4"
}

@inproceedings{efficient_biomed_rag,
   title={Efficient and Reproducible Biomedical Question Answering Using Retrieval Augmented Generation},
   url={http://dx.doi.org/10.1109/SDS66131.2025.00029},
   DOI={10.1109/sds66131.2025.00029},
   booktitle={2025 IEEE Swiss Conference on Data Science (SDS)},
   publisher={IEEE},
   author={Stuhlmann, Linus and Saxer, Michael Alexander and Fürst, Jonathan},
   year={2025},
   month={June}, pages={154–157} }

@misc{dynamic_context_selection,
      title={Dynamic Context Selection for Retrieval-Augmented Generation: Mitigating Distractors and Positional Bias}, 
      author={Malika Iratni and Mohand Boughanem and Taoufiq Dkaki},
      year={2025},
      eprint={2512.14313},
      archivePrefix={arXiv},
      primaryClass={cs.IR},
      url={https://arxiv.org/abs/2512.14313}, 
}

@misc{optimal_search_rag,
      title={Toward Optimal Search and Retrieval for RAG}, 
      author={Alexandria Leto and Cecilia Aguerrebere and Ishwar Bhati and Ted Willke and Mariano Tepper and Vy Ai Vo},
      year={2024},
      eprint={2411.07396},
      archivePrefix={arXiv},
      primaryClass={cs.CL},
      url={https://arxiv.org/abs/2411.07396}, 
}

@article{medaesqa,
  title = {A Dataset of Medical Questions Paired with Automatically Generated Answers and Evidence-supported References},
  volume = {12},
  ISSN = {2052-4463},
  url = {http://dx.doi.org/10.1038/s41597-025-05233-z},
  DOI = {10.1038/s41597-025-05233-z},
  number = {1},
  journal = {Scientific Data},
  publisher = {Springer Science and Business Media LLC},
  author = {Gupta,  Deepak and Bartels,  Davis and Demner-Fushman,  Dina},
  year = {2025},
  month = jun 
}

@misc{miriad,
      title={MIRIAD: Augmenting LLMs with millions of medical query-response pairs}, 
      author={Qinyue Zheng and Salman Abdullah and Sam Rawal and Cyril Zakka and Sophie Ostmeier and Maximilian Purk and Eduardo Reis and Eric J. Topol and Jure Leskovec and Michael Moor},
      year={2025},
      eprint={2506.06091},
      archivePrefix={arXiv},
      primaryClass={cs.CL},
      url={https://arxiv.org/abs/2506.06091}, 
}

@inproceedings{treerag,
    title = "{T}ree{RAG}: Unleashing the Power of Hierarchical Storage for Enhanced Knowledge Retrieval in Long Documents",
    author = "Tao, Wenyu  and
      Xing, Xiaofen  and
      Chen, Yirong  and
      Huang, Linyi  and
      Xu, Xiangmin",
    editor = "Che, Wanxiang  and
      Nabende, Joyce  and
      Shutova, Ekaterina  and
      Pilehvar, Mohammad Taher",
    booktitle = "Findings of the Association for Computational Linguistics: ACL 2025",
    month = jul,
    year = "2025",
    address = "Vienna, Austria",
    publisher = "Association for Computational Linguistics",
    url = "https://aclanthology.org/2025.findings-acl.20/",
    doi = "10.18653/v1/2025.findings-acl.20",
    pages = "356--371",
    ISBN = "979-8-89176-256-5"
}

@inproceedings{medrag,
    title = "Benchmarking Retrieval-Augmented Generation for Medicine",
    author = "Xiong, Guangzhi  and
      Jin, Qiao  and
      Lu, Zhiyong  and
      Zhang, Aidong",
    editor = "Ku, Lun-Wei  and
      Martins, Andre  and
      Srikumar, Vivek",
    booktitle = "Findings of the Association for Computational Linguistics: ACL 2024",
    month = aug,
    year = "2024",
    address = "Bangkok, Thailand",
    publisher = "Association for Computational Linguistics",
    url = "https://aclanthology.org/2024.findings-acl.372/",
    doi = "10.18653/v1/2024.findings-acl.372",
    pages = "6233--6251"
}

@article{bm25,
author = {Robertson, Stephen and Zaragoza, Hugo},
year = {2009},
month = {09},
pages = {333-389},
title = {The Probabilistic Relevance Framework: BM25 and Beyond},
volume = {3},
journal = {Foundations and Trends in Information Retrieval},
doi = {10.1561/1500000019}
}

@inproceedings{bge_base_emb,
author = {Xiao, Shitao and Liu, Zheng and Zhang, Peitian and Muennighoff, Niklas and Lian, Defu and Nie, Jian-Yun},
title = {C-Pack: Packed Resources For General Chinese Embeddings},
year = {2024},
isbn = {9798400704314},
publisher = {Association for Computing Machinery},
address = {New York, NY, USA},
url = {https://doi.org/10.1145/3626772.3657878},
doi = {10.1145/3626772.3657878},
booktitle = {Proceedings of the 47th International ACM SIGIR Conference on Research and Development in Information Retrieval},
pages = {641–649},
numpages = {9},
location = {Washington DC, USA},
series = {SIGIR '24}
}

@article{qwen3_emb,
  title={Qwen3 Embedding: Advancing Text Embedding and Reranking Through Foundation Models},
  author={Zhang, Yanzhao and Li, Mingxin and Long, Dingkun and Zhang, Xin and Lin, Huan and Yang, Baosong and Xie, Pengjun and Yang, An and Liu, Dayiheng and Lin, Junyang and Huang, Fei and Zhou, Jingren},
  journal={arXiv preprint arXiv:2506.05176},
  year={2025}
}

@misc{jina_rerank,
      title={jina-reranker-v3: Last but Not Late Interaction for Document Reranking}, 
      author={Feng Wang and Yuqing Li and Han Xiao},
      year={2025},
      eprint={2509.25085},
      archivePrefix={arXiv},
      primaryClass={cs.CL},
      url={https://arxiv.org/abs/2509.25085}, 
}

@misc{bge_m3_rerank1,
      title={Making Large Language Models A Better Foundation For Dense Retrieval}, 
      author={Chaofan Li and Zheng Liu and Shitao Xiao and Yingxia Shao},
      year={2023},
      eprint={2312.15503},
      archivePrefix={arXiv},
      primaryClass={cs.CL}
}

@inproceedings{bge_m3_rerank2,
    title = "{M}3-Embedding: Multi-Linguality, Multi-Functionality, Multi-Granularity Text Embeddings Through Self-Knowledge Distillation",
    author = "Chen, Jianlyu  and
      Xiao, Shitao  and
      Zhang, Peitian  and
      Luo, Kun  and
      Lian, Defu  and
      Liu, Zheng",
    editor = "Ku, Lun-Wei  and
      Martins, Andre  and
      Srikumar, Vivek",
    booktitle = "Findings of the Association for Computational Linguistics: ACL 2024",
    month = aug,
    year = "2024",
    address = "Bangkok, Thailand",
    publisher = "Association for Computational Linguistics",
    url = "https://aclanthology.org/2024.findings-acl.137/",
    doi = "10.18653/v1/2024.findings-acl.137",
    pages = "2318--2335",
}

@inproceedings{
mmlu,
title={Measuring Massive Multitask Language Understanding},
author={Dan Hendrycks and Collin Burns and Steven Basart and Andy Zou and Mantas Mazeika and Dawn Song and Jacob Steinhardt},
booktitle={International Conference on Learning Representations},
year={2021},
url={https://openreview.net/forum?id=d7KBjmI3GmQ}
}

@Article{medqa,
AUTHOR = {Jin, Di and Pan, Eileen and Oufattole, Nassim and Weng, Wei-Hung and Fang, Hanyi and Szolovits, Peter},
TITLE = {What Disease Does This Patient Have? A Large-Scale Open Domain Question Answering Dataset from Medical Exams},
JOURNAL = {Applied Sciences},
VOLUME = {11},
YEAR = {2021},
NUMBER = {14},
ARTICLE-NUMBER = {6421},
URL = {https://www.mdpi.com/2076-3417/11/14/6421},
ISSN = {2076-3417},
DOI = {10.3390/app11146421}
}

@InProceedings{medmcqa,
  title = 	 {MedMCQA: A Large-scale Multi-Subject Multi-Choice Dataset for Medical domain Question Answering},
  author =       {Pal, Ankit and Umapathi, Logesh Kumar and Sankarasubbu, Malaikannan},
  booktitle = 	 {Proceedings of the Conference on Health, Inference, and Learning},
  pages = 	 {248--260},
  year = 	 {2022},
  editor = 	 {Flores, Gerardo and Chen, George H and Pollard, Tom and Ho, Joyce C and Naumann, Tristan},
  volume = 	 {174},
  series = 	 {Proceedings of Machine Learning Research},
  month = 	 {07--08 Apr},
  publisher =    {PMLR},
  url = 	 {https://proceedings.mlr.press/v174/pal22a.html},
  }

@inproceedings{pubmedqa,
  title={PubMedQA: A Dataset for Biomedical Research Question Answering},
  author={Jin, Qiao and Dhingra, Bhuwan and Liu, Zhengping and Cohen, William and Lu, Xinghua},
  booktitle={Proceedings of the 2019 Conference on Empirical Methods in Natural Language Processing and the 9th International Joint Conference on Natural Language Processing (EMNLP-IJCNLP)},
  pages={2567--2577},
  year={2019}
}

@article{bioasq,
author = {Tsatsaronis, George and Balikas, Georgios and Malakasiotis, Prodromos and Partalas, Ioannis and Zschunke, Matthias and Alvers, Michael and Weißenborn, Dirk and Krithara, Anastasia and Petridis, Sergios and Polychronopoulos, Dimitris and Almirantis, Yannis and Pavlopoulos, John and Baskiotis, Nicolas and Gallinari, Patrick and Artieres, Thierry and Ngonga Ngomo, Axel-Cyrille and Heino, Norman and Gaussier, Eric and Barrio-Alvers, Liliana and Paliouras, Georgios},
year = {2015},
month = {04},
pages = {138},
title = {An overview of the BIOASQ large-scale biomedical semantic indexing and question answering competition},
volume = {16},
journal = {BMC Bioinformatics},
doi = {10.1186/s12859-015-0564-6}
}

@misc{gemma3,
      title={Gemma 3 Technical Report}, 
      author={Gemma Team and Aishwarya Kamath and Johan Ferret and Shreya Pathak and Nino Vieillard and Ramona Merhej and Sarah Perrin and Tatiana Matejovicova and Alexandre Ramé and Morgane Rivière and Louis Rouillard and Thomas Mesnard and Geoffrey Cideron and Jean-bastien Grill and Sabela Ramos and Edouard Yvinec and Michelle Casbon and Etienne Pot and Ivo Penchev and Gaël Liu and Francesco Visin and Kathleen Kenealy and Lucas Beyer and Xiaohai Zhai and Anton Tsitsulin and Robert Busa-Fekete and Alex Feng and Noveen Sachdeva and Benjamin Coleman and Yi Gao and Basil Mustafa and Iain Barr and Emilio Parisotto and David Tian and Matan Eyal and Colin Cherry and Jan-Thorsten Peter and Danila Sinopalnikov and Surya Bhupatiraju and Rishabh Agarwal and Mehran Kazemi and Dan Malkin and Ravin Kumar and David Vilar and Idan Brusilovsky and Jiaming Luo and Andreas Steiner and Abe Friesen and Abhanshu Sharma and Abheesht Sharma and Adi Mayrav Gilady and Adrian Goedeckemeyer and Alaa Saade and Alex Feng and Alexander Kolesnikov and Alexei Bendebury and Alvin Abdagic and Amit Vadi and András György and André Susano Pinto and Anil Das and Ankur Bapna and Antoine Miech and Antoine Yang and Antonia Paterson and Ashish Shenoy and Ayan Chakrabarti and Bilal Piot and Bo Wu and Bobak Shahriari and Bryce Petrini and Charlie Chen and Charline Le Lan and Christopher A. Choquette-Choo and CJ Carey and Cormac Brick and Daniel Deutsch and Danielle Eisenbud and Dee Cattle and Derek Cheng and Dimitris Paparas and Divyashree Shivakumar Sreepathihalli and Doug Reid and Dustin Tran and Dustin Zelle and Eric Noland and Erwin Huizenga and Eugene Kharitonov and Frederick Liu and Gagik Amirkhanyan and Glenn Cameron and Hadi Hashemi and Hanna Klimczak-Plucińska and Harman Singh and Harsh Mehta and Harshal Tushar Lehri and Hussein Hazimeh and Ian Ballantyne and Idan Szpektor and Ivan Nardini and Jean Pouget-Abadie and Jetha Chan and Joe Stanton and John Wieting and Jonathan Lai and Jordi Orbay and Joseph Fernandez and Josh Newlan and Ju-yeong Ji and Jyotinder Singh and Kat Black and Kathy Yu and Kevin Hui and Kiran Vodrahalli and Klaus Greff and Linhai Qiu and Marcella Valentine and Marina Coelho and Marvin Ritter and Matt Hoffman and Matthew Watson and Mayank Chaturvedi and Michael Moynihan and Min Ma and Nabila Babar and Natasha Noy and Nathan Byrd and Nick Roy and Nikola Momchev and Nilay Chauhan and Noveen Sachdeva and Oskar Bunyan and Pankil Botarda and Paul Caron and Paul Kishan Rubenstein and Phil Culliton and Philipp Schmid and Pier Giuseppe Sessa and Pingmei Xu and Piotr Stanczyk and Pouya Tafti and Rakesh Shivanna and Renjie Wu and Renke Pan and Reza Rokni and Rob Willoughby and Rohith Vallu and Ryan Mullins and Sammy Jerome and Sara Smoot and Sertan Girgin and Shariq Iqbal and Shashir Reddy and Shruti Sheth and Siim Põder and Sijal Bhatnagar and Sindhu Raghuram Panyam and Sivan Eiger and Susan Zhang and Tianqi Liu and Trevor Yacovone and Tyler Liechty and Uday Kalra and Utku Evci and Vedant Misra and Vincent Roseberry and Vlad Feinberg and Vlad Kolesnikov and Woohyun Han and Woosuk Kwon and Xi Chen and Yinlam Chow and Yuvein Zhu and Zichuan Wei and Zoltan Egyed and Victor Cotruta and Minh Giang and Phoebe Kirk and Anand Rao and Kat Black and Nabila Babar and Jessica Lo and Erica Moreira and Luiz Gustavo Martins and Omar Sanseviero and Lucas Gonzalez and Zach Gleicher and Tris Warkentin and Vahab Mirrokni and Evan Senter and Eli Collins and Joelle Barral and Zoubin Ghahramani and Raia Hadsell and Yossi Matias and D. Sculley and Slav Petrov and Noah Fiedel and Noam Shazeer and Oriol Vinyals and Jeff Dean and Demis Hassabis and Koray Kavukcuoglu and Clement Farabet and Elena Buchatskaya and Jean-Baptiste Alayrac and Rohan Anil and Dmitry and Lepikhin and Sebastian Borgeaud and Olivier Bachem and Armand Joulin and Alek Andreev and Cassidy Hardin and Robert Dadashi and Léonard Hussenot},
      year={2025},
      eprint={2503.19786},
      archivePrefix={arXiv},
      primaryClass={cs.CL},
      url={https://arxiv.org/abs/2503.19786}, 
}

@article{medgemma1.5,
  title={MedGemma 1.5 Technical Report},
  author={Sellergren, Andrew and Gao, Chufan and Mahvar, Fereshteh and Kohlberger, Timo and Jamil, Fayaz and Traverse, Madeleine and Tono, Alberto and Sadjad, Bashir and Yang, Lin and Lau, Charles and others},
  journal={arXiv preprint arXiv:2604.05081},
  year={2026}
}

@misc{qwen3.5,
  author       = {{Qwen Team}},
  title        = {{Qwen3.5-4B}},
  year         = {2026},
  howpublished = {Hugging Face model card},
  url          = {https://huggingface.co/Qwen/Qwen3.5-4B},
  note         = {Retrieved on September 3, 2026}
}

\makeatletter
\edef\articlelastpage{\thepage}
\AtEndDocument{\immediate\write\@auxout{\string\csxdef{lastpage}{\articlelastpage}}}
\makeatother
\clearpage
\onecolumn


\section*{Supplementary material}

# Supplementary Material

## Large-Scale Biomedical RAG: Corpus Scaling, Document Representation, and Efficiency Trade-offs

Adrián Sánchez-Freire, Gonzalo Pérez-Gómez, Aitziber Atutxa and Ander Barrena

*HiTZ Basque Center for Language Technology,*
*University of the Basque Country (EHU), Spain*

---

**Supplementary file description.** This document provides supplementary implementation and reproducibility details, complete development-set results, and the prompt templates used in the experimental pipeline. Main sections use appendix letters (A, B, C, . . . ), and tables, figures, and equations are numbered within their corresponding section (e.g., Table A.1 and Figure F.1).

---

## Contents

# A Implementation Details and Reproducibility

## A.1 Computational Environment

Experiments were executed on a shared SLURM-managed compute environment. The CPU model used for index construction and retrieval experiments was an `AMD EPYC 7513 32-Core Processor`. Indexing jobs were submitted with 6 SLURM-allocated CPU cores. GPU-accelerated stages were configured to use up to four `NVIDIA A100-SXM4-80GB` GPUs in parallel, with the effective number of GPUs per job varying between one and four depending on job placement, resource availability, and workload characteristics. The operating system was Linux kernel 4.18.0 with glibc 2.28.

## A.2 Software, Libraries, and Versions

Table A.1 reports the main software libraries and versions used in the experiments.

| Library | Version |
|---|---|
| `numpy` | 2.2.6 |
| `pandas` | 3.0.0 |
| `torch` | 2.10.0 |
| `transformers` | 5.8.0 |
| `sentence-transformers` | 5.4.1 |
| `vllm` | 0.19.1 |
| `datasets` | 4.5.0 |
| `elasticsearch` | 8.19.3 |
| `pyserini` | 1.5.0 |
| `faiss-cpu` | 1.13.2 |
| `lancedb` | 0.29.2 |

| Library | Version |
|---|---|
| `langchain-core` | 1.2.19 |
| `langchain-community` | 0.4.1 |
| `langchain-text-splitters` | 1.1.1 |
| `pyyaml` | 6.0.3 |
| `orjson` | 3.11.7 |
| `tqdm` | 4.67.3 |
| `psutil` | 7.2.2 |
| `codecarbon` | 3.2.3 |
| `matplotlib` | 3.10.8 |
| `openpyxl` | 3.1.5 |

Table A.1: Main software libraries and versions used in the experiments.

Model inference was performed with either `vLLM` or the Hugging Face `transformers` stack depending on the experimental stage. `vLLM` was used for high-throughput generation and embedding extraction during dataset conversion, pseudo-abstract generation, and index construction. The reported retriever, reranker, and end-to-end MCQA evaluations were performed with Hugging Face-based implementations.

Elasticsearch 8.19.3 was used as the backend for the sparse and dense retrieval indexes. Sparse retrieval used BM25 over the indexed text field, while dense retrieval used indexed `dense_vector` fields with cosine similarity and `bbq_hnsw` approximate nearest-neighbor search. Indexes were built once before evaluation and reused across the retriever and reranker experiments.

# B MCQA Conversion Implementation Details

This section specifies the implementation details of the open-ended to MCQA conversion procedure described in the main manuscript. Prompt templates are provided in Supplementary Section F.

The conversion pipeline is applied independently to each MedAESQA and MIRIAD question–answer pair. Each instance preserves the original reference answer as the gold option and produces three incorrect distractors, resulting in a four-option MCQA item with a single answer key. The pipeline consists of raw distractor generation, distractor rewriting, distractor filtering and ranking, MCQA construction, and answer-position balancing.

## B.1 Raw Distractor Generation

For each question, we first generate a pool of candidate distractors from the original question and reference answer. An instruction-tuned language model is prompted to produce exactly ten medically plausible but incorrect distractors. The prompt requires candidates to be non-overlapping

with the correct answer, diverse, non-redundant, and similar in length to the reference answer. The output is constrained to a JSON array to facilitate automatic parsing.

Generation is performed using `meta-llama/Llama-3.3-70B-Instruct` with sampling enabled, temperature 0.7, top-$p$ 0.95, and a maximum generation length of 1000 new tokens. The prompt template is shown in Figure F.1.

### B.2 Distractor Rewriting and Length Calibration

Because raw distractors may introduce superficial cues, especially through differences in length or syntactic structure, each candidate distractor is passed through a rewriting stage. The model receives the original question, the reference answer, and the draft distractor, and is instructed to preserve the distractor concept while matching the reference answer in syntactic structure, word length, technical density, and visual footprint.

The generated text is post-processed to remove surrounding quotes and common conversational prefixes. Empty outputs are discarded by falling back to the original distractor. Candidate distractors that exactly match the reference answer are skipped before rewriting.

Rewriting is performed using `meta-llama/Llama-3.3-70B-Instruct` with sampling enabled, temperature 0.7, top-$p$ 0.95, and a maximum generation length of 400 new tokens. The prompt template is shown in Figure F.2.

### B.3 Distractor Filtering, Ranking, and MCQA Construction

After rewriting, candidate distractors are cleaned by removing empty or malformed outputs, exact duplicates, and candidates that are nearly identical to the correct answer. Near-identity is detected using a two-step heuristic: a candidate is removed only when its character-level similarity to the correct answer is above 0.85 and the two strings differ by at most one word. This avoids discarding candidates that share surface form with the correct answer but differ substantially in content.

If fewer than three valid rewritten distractors remain, the pipeline falls back to the original raw distractors and repeats the cleaning procedure. The remaining candidates are then ranked by an instruction-tuned model, which selects the three best distractors according to medical plausibility, non-redundancy, and lack of semantic overlap with the correct answer.

Ranking is performed using `meta-llama/Llama-3.3-70B-Instruct` with deterministic decoding and a maximum generation length of 200 new tokens. The model outputs the indices of the selected distractors, which are combined with the reference answer to form the final four-option MCQA instance. The prompt template is shown in Figure F.3.

### B.4 Answer Position Balancing

As a final post-processing step, we balance the position of the correct answer across labels A, B, C, and D. The converted dataset is first constructed without enforcing a uniform answer-key distribution. We then define an approximately balanced pool of target answer labels, shuffle it, and assign one target label to each question. For each instance, the correct answer is moved to the assigned position, and the three distractors are randomly placed in the remaining slots.

This preserves the content of all answer options while reducing positional bias in the final MCQA dataset.

## C MIRIAD Pseudo-Abstract Generation Details

This section specifies the implementation details of the pseudo-abstract generation procedure described in the main manuscript. The prompt template is provided in Supplementary Section F.

Because MIRIAD does not provide separate abstracts, we generate pseudo-abstracts from the original MIRIAD passages to define an abstract-level representation analogous to the PubMed

abstract setting. The original passages are still used directly for the full-text chunking representation and as source documents for TreeRAG construction.

Pseudo-abstracts are generated independently for each passage using `Qwen/Qwen3.5-35B-A3B` with vLLM. Passages containing 300 tokens or fewer are not summarized and are retained unchanged, since they already match the target granularity of the abstract-level representation.

Generation is performed with thinking mode disabled, temperature 0.2, top-$p$ 0.95, a maximum generation length of 450 tokens, and repetition penalty 1.1. The prompt template is shown in Figure F.4.

For the abstract-level representation, the resulting pseudo-abstract is stored in the `content` field. As for the other document representations, the indexed `contents` field is constructed by concatenating the title, when available, with the corresponding `content` field.

# D TreeRAG Construction Details

This section describes the implementation details of the TreeRAG-style hierarchical representation described in the main manuscript. The prompt template used for hierarchical segmentation is provided in Supplementary Section F.

For each source document, the text is first split into paragraphs, and each paragraph is assigned a stable integer identifier. The generative model is not asked to rewrite or regenerate the document text. Instead, it outputs a structured hierarchy containing node identifiers, short synthetic titles, and lists of paragraph identifiers associated with each node. This design keeps the final indexed content grounded in the original document text and reduces the risk of introducing generated content into the retrieval units.

The hierarchy is generated dynamically for each document. Therefore, the number of nodes and the depth of the tree are not fixed in advance, but depend on the thematic structure of the source document. Each node groups semantically related paragraphs and receives a short synthetic title summarizing its local content. For documents that exceed the model context window, the input is processed in smaller consecutive blocks before reconstructing the final hierarchy.

TreeRAG construction was performed using `Qwen/Qwen3.5-35B-A3B` with `vLLM`. Generation was run with thinking mode disabled, temperature 0.1, top-$p$ 0.8, and a maximum generation length of 8000 tokens. The model output was parsed automatically to extract node identifiers, synthetic titles, and paragraph indices. Malformed nodes and invalid paragraph references were removed during post-processing.

After parsing, paragraph identifiers are replaced with the corresponding original paragraph text. Each retrieval unit is then built by concatenating the synthetic titles inherited from its ancestor nodes with the original text associated with the current node. The resulting hierarchy-enriched snippets are indexed using the same retrieval backend as the other document representations.

Because TreeRAG requires an additional generative preprocessing step, its construction cost is substantially higher than flat full-text chunking. This makes TreeRAG more expensive to build at scale, although the resulting retrieval units can still be indexed and queried using the same backend as the other representations.

# E Phase III End-to-End Details

This section provides the detailed end-to-end MCQA accuracy results for the DEV datasets used in Phase III, together with the decoding configuration used by the generator models. It complements the aggregate trends shown in the Phase III figures of the main manuscript by reporting absolute accuracy values for each generator, corpus scale, document representation, and evaluation setting.

All generator models were evaluated with deterministic decoding, using greedy generation and a repetition penalty of 1.1. The maximum number of generated tokens was set to 100 for standard MCQA generators and to 16,384 for reasoning-oriented generators. Because the decoding process is deterministic, re-running the same generator–retrieval configuration with the same prompts does not

introduce sampling-induced variability. We therefore report one accuracy value per configuration rather than averages over multiple stochastic runs.

This setting is appropriate for MCQA evaluation because the goal is to measure the accuracy of each fixed generator–retrieval configuration under a controlled decoding policy, rather than to estimate variability induced by sampling. Results for MedAESQA and MIRIAD DEV are reported in Tables E.1 and E.2, respectively.

| Generator | Scale | LLM-only | RAG | | | Oracle |
|---|---|---|---|---|---|---|
| | | | Abstract | Full-text | TreeRAG | |
| Gemma-3 | Control | 65.0 | 87.5 | 80.0 | **90.0** | 92.5 |
| | Tiny | 65.0 | 87.5 | 80.0 | **92.5** | |
| | Small | 65.0 | 87.5 | 80.0 | **90.0** | |
| | Medium | 65.0 | 85.0 | 77.5 | **90.0** | |
| | Large | 65.0 | **87.5** | 80.0 | 80.0 | |
| | Full | 65.0 | **85.0** | 75.0 | 77.5 | |
| MedGemma-1.5 | Control | 60.0 | **80.0** | 60.0 | 75.0 | 80.0 |
| | Tiny | 60.0 | **77.5** | 62.5 | 75.0 | |
| | Small | 60.0 | **77.5** | 60.0 | **77.5** | |
| | Medium | 60.0 | 67.5 | 60.0 | **75.0** | |
| | Large | 60.0 | 72.5 | 60.0 | **80.0** | |
| | Full | 60.0 | 62.5 | **67.5** | 62.5 | |
| Qwen3.5 | Control | 75.0 | **95.0** | 85.0 | 90.0 | 97.5 |
| | Tiny | 75.0 | **95.0** | 85.0 | 90.0 | |
| | Small | 75.0 | **95.0** | 85.0 | 90.0 | |
| | Medium | 75.0 | **95.0** | 85.0 | 87.5 | |
| | Large | 75.0 | **92.5** | 85.0 | 87.5 | |
| | Full | 75.0 | **92.5** | 80.0 | 85.0 | |
| Qwen3.5-Think | Control | 85.0 | **97.5** | 90.0 | **97.5** | 100.0 |
| | Tiny | 85.0 | **97.5** | 90.0 | 92.5 | |
| | Small | 85.0 | **95.0** | 87.5 | **95.0** | |
| | Medium | 85.0 | **95.0** | 90.0 | 90.0 | |
| | Large | 85.0 | **92.5** | 87.5 | **92.5** | |
| | Full | 85.0 | **90.0** | 85.0 | 85.0 | |

Table E.1: End-to-end MCQA accuracy (%) on MedAESQA for each generator across corpus scales and RAG document representations. All generators correspond to their 4B versions. Bold values indicate the best RAG result for each generator and scale.

| Generator | Scale | LLM-only | RAG | | | Oracle |
|---|---|---|---|---|---|---|
| | | | Abstract | Full-text | TreeRAG | |
| Gemma-3 | Control | 67.5 | 90.5 | 99.0 | **99.5** | 98.5 |
| | Tiny | 67.5 | 90.5 | 99.0 | **99.5** | |
| | Small | 67.5 | 91.0 | **98.5** | **98.5** | |
| | Medium | 67.5 | 90.5 | 95.0 | **97.0** | |
| | Large | 67.5 | 87.0 | **97.0** | 96.0 | |
| MedGemma-1.5 | Control | 65.0 | 79.5 | 93.0 | **94.5** | 93.5 |
| | Tiny | 65.0 | 79.5 | 93.0 | **95.5** | |
| | Small | 65.0 | 83.0 | 91.5 | **93.5** | |
| | Medium | 65.0 | 81.0 | 91.0 | **93.0** | |
| | Large | 65.0 | 77.5 | **90.5** | **90.5** | |
| Qwen3.5 | Control | 85.0 | 97.5 | **100.0** | **100.0** | 100.0 |
| | Tiny | 85.0 | 96.0 | **100.0** | **100.0** | |
| | Small | 85.0 | 97.5 | **100.0** | **100.0** | |
| | Medium | 85.0 | 95.0 | 98.5 | **99.5** | |
| | Large | 85.0 | 92.0 | **98.5** | 97.0 | |
| Qwen3.5-Think | Control | 85.0 | 97.0 | **100.0** | **100.0** | 100.0 |
| | Tiny | 85.0 | 97.0 | **100.0** | **100.0** | |
| | Small | 85.0 | 95.5 | **99.5** | **99.5** | |
| | Medium | 85.0 | 95.5 | **99.0** | **99.0** | |
| | Large | 85.0 | 93.5 | 97.5 | **98.0** | |

Table E.2: End-to-end MCQA accuracy (%) on MIRIAD DEV for each generator across corpus scales and RAG document representations. All generators correspond to their 4B versions. Bold values indicate the best RAG result for each generator and scale.

# F Prompt Templates

This section reports the complete prompt templates used throughout the experimental pipeline. Figures F.1, F.2, and F.3 correspond to the three prompts used for MCQA conversion: raw distractor generation, distractor rewriting, and final distractor selection. Figures F.4, F.5, and F.6 show the prompts used for MIRIAD pseudo-abstract generation, hierarchical segmentation during TreeRAG construction, and end-to-end MCQA answering, respectively.

In the end-to-end prompt, the allowed output letters are A, B, C, D, and E. Although the datasets evaluated in this work contain at most four options, and some contain only two or three, the broader answer format keeps the same prompt template transferable to MCQA datasets with different numbers of answer choices.

**Prompt template for raw distractor generation**

### System prompt

```
You are an expert medical educator who generates plausible but incorrect multiple-choice
distractors.

Rules:
- Distractors must be medically plausible but clearly incorrect.
- They must not overlap with or imply the correct answer.
- They must be SIMILAR IN LENGTH to the correct answer.
- They must be diverse and non-redundant.
- You MUST output ONLY a valid JSON array of exactly 10 strings.
- Do NOT output explanations or commentary.
- Do NOT output anything before or after the JSON.
```

---

### User prompt

```
Question:
{question}

Correct answer:
{answer}

TASK:
Generate exactly 10 high-quality incorrect distractors that follow all rules.

OUTPUT FORMAT (strict):
[
  "distractor one",
  "distractor two",
  "..."
]
Return ONLY the JSON array.
```

Figure F.1: Prompt template used to generate candidate distractors from each open-ended question–answer pair.

**Prompt template for distractor expansion**

**System prompt**

```
You are a medical psychometrician specialized in eliminating 'length cue' bias in board-style
exams. Your goal is to make the Distractor indistinguishable from the Reference Answer by
mirroring its syntactic structure, technical density, and visual footprint. You prioritize
structural symmetry over creative expansion.
```

---

**User prompt**

```
### CONTEXT
The current distractor creates length bias. You must recalibrate it to match the Reference
Answer's profile.

### INPUT DATA
**Question:** "{question}"
**Reference Answer (The Pattern):** "{answer}"
**Draft Distractor (The Concept):** "{distractor}"

### INSTRUCTIONS
1. SYNTACTIC MIRRORING: Identify the grammatical structure of the Reference Answer (e.g.,
[Mechanism] leading to [Effect]). Rewrite the Distractor using this exact same template.

2. LENGTH CALIBRATION: The output must be within +/- 3 words of the Reference Answer. Do not
exceed this range.

3. TECHNICAL DENSITY: Match the number of specific medical entities (enzymes, receptors,
anatomical sites) used in the Reference Answer. Use high-level terminology to reach the
target length rather than simple adjectives.

4. FACTUALITY: Ensure the final text remains medically incorrect for the given question.

5. OUTPUT RULE: Provide ONLY the final rewritten text. No quotes, no preamble, no
explanations.

### REWRITTEN DISTRACTOR:
```

Figure F.2: Prompt template used to rewrite candidate distractors and reduce length-cue bias.

**Prompt template for distractor ranking**

### System prompt

```
You are an expert evaluator of medical MCQ distractors.
You will be provided with a Question, the Correct Answer, and a numbered list of Candidate
Distractors.

Your task is to identify the INDICES (numbers) of the 3 best distractors.

Criteria for best distractors:
1. Medically plausible but incorrect.
2. Not redundant with each other.
3. Not a rephrasing of the correct answer.

OUTPUT FORMAT:
Return ONLY a JSON object with the list of indices (0-based integers).
Example: {"best_indices": [0, 4, 7]}
```

---

### User prompt

```
Question:
{question}

Correct Answer:
{answer}

Candidate Distractors:
{distractors}

TASK:
Return the JSON object with the indices of the 3 best distractors.
```

Figure F.3: Prompt template used to select the final three distractors from the candidate pool.

**Prompt template for MIRIAD pseudo-abstract generation**

**System prompt**

```
You are a biomedical summarization assistant.
```

---

**User prompt**

```
Task: Write a concise abstract-style summary of the following passage.

Requirements:
- 4-6 sentences.
- Target length: approximately 230-300 words.
- Use formal, academic medical language.
- Structure: briefly introduce the context/objective, describe key methods or findings, and
state the main conclusion.
- Do NOT include greetings, explanations, or meta-commentary. Output only the summary.

Passage:
{content}

Summary:
```

Figure F.4: Prompt template used to generate pseudo-abstracts for MIRIAD.

**Prompt template for hierarchical medical article chunking**

**System prompt**

```
You are a Medical Information Specialist. Your goal is to structure medical articles by
grouping paragraph indices into semantically coherent hierarchical chunks.
```

---

**User prompt**

```
Input text is pre-processed with paragraph indices (<n>) at the beginning of each line.

### 1. Chunking logic
- Merge only consecutive paragraphs with the same medical concept.
- Split when topic changes (e.g., Methods -> Results).

### 2. Output format
- ID: hierarchical numbering (e.g., 1, 1.1, 1.1.1)
- TITLE: short descriptive title (3-6 words)
- INDICES: list of paragraph indices
- NO TEXT: output only structure

### Example

ID: 1
TITLE: Clinical Trial Design
INDICES: [<1>, <2>, <3>]

Input:
{document}
```

Figure F.5: Prompt template used for hierarchical chunking of medical articles. The variable is `document`.

**Prompt template for end-to-end MCQA generation**

### System prompt

```
You are a multiple-choice question answering assistant.
If context is provided, use the information from it to choose the best answer.
If no context is provided, answer using your own knowledge.

Answer format rules:
- Respond with ONLY one letter: A, B, C, D, or E.
- Do NOT include explanations or extra text.
```

---

### User prompt

```
Context:
{context}
---
Question: {question}

choices:
{choices}

Answer:
```

Figure F.6: Prompt template used for end-to-end multiple-choice answer generation. The variables are `context`, `question`, and `choices`.



\end{document}